\documentclass[11pt]{article}
\usepackage[preprint]{acl}
\usepackage{times}
\usepackage{latexsym}
\usepackage[T1]{fontenc}
\usepackage[utf8]{inputenc}
\usepackage{microtype}
\usepackage{tgheros}

\usepackage{amsmath,amssymb,amsthm}
\usepackage{graphicx}
\usepackage{booktabs,multirow,array}
\usepackage{tabularx}
\usepackage{tikz}
\usetikzlibrary{arrows.meta,positioning,fit,backgrounds,calc}
\usepackage{xcolor}
\usepackage{enumitem}
\usepackage{caption}
\theoremstyle{definition}
\newtheorem{definition}{Definition}
\newtheorem{assumption}{Assumption}
\theoremstyle{plain}
\newtheorem{proposition}{Proposition}
\newtheorem{identity}{Identity}

\newcommand{\ALGem}{$+$0.35}

\newcommand{\ALMis}{$+$3.11}

\newcommand{\ALQwe}{$-$0.03}

\newcommand{\APubGem}{$-$2.19}

\newcommand{\AllThree}{43}
\newcommand{\ArmIfourB}{18}

\newcommand{\ArmIfourV}{40}

\newcommand{\ArmTfourB}{2}

\newcommand{\ArmTfourV}{40}

\newcommand{\AthreeLoopB}{16}
\newcommand{\AthreeLoopN}{327}
\newcommand{\AthreeLoopRate}{0.05}
\newcommand{\AthreeProbe}{290/290}

\newcommand{\CachedN}{17}
\newcommand{\CachedOwn}{16}
\newcommand{\CachedTot}{120}

\newcommand{\CatDiffQwe}{$+$0.18}
\newcommand{\CatDiffQweCI}{[0.02, 0.34]}

\newcommand{\CatStratQweCI}{[0.06, 0.30]}

\newcommand{\ConsFTA}{29}

\newcommand{\ConsNeg}{2}

\newcommand{\CostHidGem}{47}
\newcommand{\CostHidMis}{11}
\newcommand{\CostHidQwe}{40}
\newcommand{\CostPctGem}{32}
\newcommand{\CostPctMax}{63}
\newcommand{\CostPctMin}{28}
\newcommand{\CostPctMis}{63}
\newcommand{\CostPctQwe}{28}
\newcommand{\CostRedGem}{103}
\newcommand{\CostRedMis}{226}
\newcommand{\CostRedQwe}{98}
\newcommand{\CostValGem}{324}
\newcommand{\CostValMis}{357}
\newcommand{\CostValQwe}{350}

\newcommand{\CrossCI}{[$-$0.73, $-$0.35]}
\newcommand{\CrossDiff}{$-$0.54}
\newcommand{\CrossFromIns}{$-$0.62}
\newcommand{\CrossFromThk}{$-$0.46}

\newcommand{\CrossPaired}{26}

\newcommand{\DeltaGem}{0.19}
\newcommand{\DeltaGemCI}{[0.13, 0.26]}
\newcommand{\DeltaInvOne}{0.36}
\newcommand{\DeltaInvOneCI}{[0.31, 0.40]}
\newcommand{\DeltaInvZero}{0.31}
\newcommand{\DeltaInvZeroCI}{[0.26, 0.36]}
\newcommand{\DeltaMacro}{0.32}
\newcommand{\DeltaMacroCI}{[0.27, 0.37]}

\newcommand{\DeltaMis}{0.58}
\newcommand{\DeltaMisCI}{[0.46, 0.67]}
\newcommand{\DeltaQwe}{0.19}
\newcommand{\DeltaQweCI}{[0.11, 0.27]}

\newcommand{\DfbShare}{47}
\newcommand{\DfbStates}{95}

\newcommand{\Down}{$-$0.20}
\newcommand{\DownCI}{[$-$0.32, $-$0.09]}
\newcommand{\DownN}{25}
\newcommand{\DstrictN}{1}

\newcommand{\FTAvBD}{16}
\newcommand{\FTAvBG}{21}
\newcommand{\GamGem}{$+$0.09}
\newcommand{\GamGemCI}{[$-$0.46, 1.12]}
\newcommand{\GamMis}{$+$0.33}
\newcommand{\GamMisCI}{[0.02, 0.89]}
\newcommand{\GamQwe}{$-$0.78}
\newcommand{\GamQweCI}{[$-$1.28, 0.26]}
\newcommand{\HBdBCI}{[$-$0.27, $-$0.11]}
\newcommand{\HBdCCI}{[$-$0.04, 0.12]}
\newcommand{\HBdDCI}{[$-$0.21, $-$0.02]}
\newcommand{\HBdECI}{[$-$0.39, $-$0.22]}
\newcommand{\HBdENCI}{[$-$0.31, $-$0.16]}
\newcommand{\HBdMCI}{[$-$0.12, 0.02]}
\newcommand{\HBdNCI}{[$-$0.14, 0.01]}
\newcommand{\HBoBCI}{[$-$3.10, $-$1.17]}
\newcommand{\HBoCCI}{[$-$0.28, 0.90]}
\newcommand{\HBoDCI}{[$-$1.98, $-$0.28]}
\newcommand{\HBoECI}{[$-$5.07, $-$2.19]}
\newcommand{\HBoENCI}{[$-$4.64, $-$1.83]}
\newcommand{\HBoMCI}{[$-$0.94, 0.28]}
\newcommand{\HBoNCI}{[$-$1.12, 0.15]}
\newcommand{\HBruns}{36}

\newcommand{\HarnRepPtwo}{1.3}
\newcommand{\HarnRepRfour}{0.5}

\newcommand{\Hfoura}{$-$0.40}
\newcommand{\HfouraCI}{[$-$0.55, $-$0.25]}

\newcommand{\Hfourb}{$+$0.06}
\newcommand{\HfourbCI}{[$-$0.06, 0.17]}

\newcommand{\HoneB}{6}

\newcommand{\Htwoa}{$-$0.19}
\newcommand{\HtwoaCI}{[$-$0.25, $-$0.13]}

\newcommand{\Htwob}{$-$0.30}
\newcommand{\HtwobCI}{[$-$0.36, $-$0.24]}

\newcommand{\Htwod}{$-$0.11}

\newcommand{\HtwodP}{0.002}

\newcommand{\INSb}{292}
\newcommand{\INSn}{997}
\newcommand{\INSrate}{0.29}
\newcommand{\InvHarnMis}{3}
\newcommand{\InvIns}{10}

\newcommand{\InvThk}{0}
\newcommand{\InvTotal}{49}
\newcommand{\InvTruncGem}{35}
\newcommand{\InvTruncQwe}{10}
\newcommand{\Kappa}{0.50}
\newcommand{\KappaAD}{0.80}

\newcommand{\KappaBlind}{0.62}

\newcommand{\LoBMis}{$-$2.02}
\newcommand{\LoBMisCI}{[$-$3.63, $-$1.09]}

\newcommand{\MisDIAB}{26}
\newcommand{\MisDIAN}{33}
\newcommand{\MisProbeB}{5}
\newcommand{\MisProbeN}{60}
\newcommand{\MisVenAB}{47}
\newcommand{\MisVenAN}{85}
\newcommand{\MisVeniceMax}{87}
\newcommand{\MisVeniceMin}{71}

\newcommand{\NMfail}{0.055}
\newcommand{\NMfailB}{254}
\newcommand{\NMfailHi}{24}
\newcommand{\NMfailN}{4{,}626}
\newcommand{\NMnear}{0.095}
\newcommand{\NMnearB}{467}
\newcommand{\NMnearHi}{23}
\newcommand{\NMnearN}{4{,}921}
\newcommand{\NMruns}{57}

\newcommand{\NxF}{6.5}
\newcommand{\NxW}{4.6}

\newcommand{\PCSGem}{24}
\newcommand{\PCSGemN}{56}
\newcommand{\PCSMis}{147}
\newcommand{\PCSMisN}{215}
\newcommand{\PCSQwe}{27}
\newcommand{\PCSQweN}{58}

\newcommand{\ProvMainN}{5}

\newcommand{\QweCatAdj}{$+$0.18}
\newcommand{\QweCatAdjCI}{[0.05, 0.32]}

\newcommand{\RPctl}{67}
\newcommand{\RPdnp}{8}

\newcommand{\RPn}{119}
\newcommand{\RPnofail}{44}

\newcommand{\RTvBD}{2}
\newcommand{\RTvBG}{3}

\newcommand{\RaFGem}{$-$1.25}

\newcommand{\RaFMis}{$+$0.81}

\newcommand{\RaFQwe}{$-$3.49}

\newcommand{\RaHGem}{$+$2.72}

\newcommand{\RaHMis}{$+$1.28}

\newcommand{\RaHQwe}{$+$2.11}

\newcommand{\ReasonN}{40}

\newcommand{\RfAH}{$-$0.21}
\newcommand{\RfAHCI}{[$-$0.30, $-$0.11]}

\newcommand{\RfAHGem}{$-$0.57}

\newcommand{\RfAHQwe}{$-$0.36}

\newcommand{\RfBa}{$-$0.17}
\newcommand{\RfBaCI}{[$-$0.23, $-$0.10]}

\newcommand{\RfBaGem}{$+$0.03}

\newcommand{\RfBaMis}{$-$0.31}

\newcommand{\RfCalls}{5{,}754}
\newcommand{\RfChanged}{$-$0.057}
\newcommand{\RfChangedCI}{[$-$0.097, $-$0.016]}
\newcommand{\RfChangedCIh}{[$-$0.113, 0.003]}
\newcommand{\RfChangedN}{243}
\newcommand{\RfEa}{$-$0.27}

\newcommand{\RfEq}{$-$0.04}
\newcommand{\RfEqA}{0.34}
\newcommand{\RfEqCI}{[$-$0.07, $-$0.01]}
\newcommand{\RfEqCIh}{[$-$0.08, 0.00]}
\newcommand{\RfEqM}{0.30}

\newcommand{\RfFN}{$-$0.16}
\newcommand{\RfFNCI}{[$-$0.22, $-$0.09]}

\newcommand{\RfFNGem}{$-$0.14}

\newcommand{\RfFNLo}{$-$1.44}

\newcommand{\RfFNMis}{$-$0.22}

\newcommand{\RfFNQwe}{$-$0.11}

\newcommand{\RfFolA}{0.03}

\newcommand{\RfFolP}{0.425}
\newcommand{\RfFolS}{0.05}

\newcommand{\RfJK}{$+$0.07}
\newcommand{\RfJKCI}{[$-$0.02, 0.16]}

\newcommand{\RfJa}{$-$0.05}

\newcommand{\RfKa}{$-$0.00}

\newcommand{\RfNA}{$-$0.06}
\newcommand{\RfNACI}{[$-$0.13, 0.00]}

\newcommand{\RfPF}{$-$0.05}
\newcommand{\RfPFCI}{[$-$0.10, $-$0.01]}

\newcommand{\RfPFGem}{$+$0.00}

\newcommand{\RfPFLo}{$-$1.15}
\newcommand{\RfPFLoCI}{[$-$2.40, $-$0.31]}

\newcommand{\RfPFMis}{$-$0.15}

\newcommand{\RfPFP}{0.054}
\newcommand{\RfPFQwe}{$-$0.01}

\newcommand{\RfRB}{$+$0.10}
\newcommand{\RfRBCI}{[0.05, 0.14]}

\newcommand{\RfRa}{$-$0.07}
\newcommand{\RfRaCI}{[$-$0.13, $-$0.01]}

\newcommand{\RfRaGem}{$+$0.05}

\newcommand{\RfRaMis}{$-$0.04}

\newcommand{\RfRaP}{0.071}
\newcommand{\RfRaQwe}{$-$0.21}

\newcommand{\RfSameN}{117}

\newcommand{\RfTa}{$-$0.10}
\newcommand{\RfTaCI}{[$-$0.33, 0.10]}

\newcommand{\RfWithT}{13}
\newcommand{\Rg}{$-$0.33}
\newcommand{\RgCI}{[$-$0.43, $-$0.22]}
\newcommand{\RgI}{0.38}

\newcommand{\RgProv}{Alibaba}
\newcommand{\RgT}{0.06}
\newcommand{\RgTokI}{107}
\newcommand{\RgTokT}{3896}

\newcommand{\RpAHMis}{$+$0.31}
\newcommand{\RpAHMisCI}{[0.12, 0.49]}

\newcommand{\RrA}{0.30}

\newcommand{\RrF}{0.09}

\newcommand{\RrP}{0.03}

\newcommand{\RthreeCalls}{179}
\newcommand{\RthreeModels}{Gemma-4-26B-A4B}
\newcommand{\RthreePlanned}{2{,}484}
\newcommand{\RthreeStates}{11}
\newcommand{\SBeight}{$-$0.14}
\newcommand{\SBeightCI}{[$-$0.24, $-$0.05]}
\newcommand{\SBthree}{$-$0.13}

\newcommand{\SEeight}{$-$0.28}
\newcommand{\SEeightCI}{[$-$0.38, $-$0.18]}
\newcommand{\SEthree}{$-$0.29}

\newcommand{\SeedAgreeMax}{0.38}
\newcommand{\SeedAgreeMin}{0.25}
\newcommand{\SeedDiff}{0.54}

\newcommand{\SeedSame}{0.44}

\newcommand{\SpentRthree}{0.029}

\newcommand{\SpentTotal}{4.88}
\newcommand{\SsEExGem}{$+$0.07}
\newcommand{\SsEExGemCI}{[$-$1.31, 1.61]}
\newcommand{\SsEExMis}{$-$1.62}
\newcommand{\SsEExMisCI}{[$-$3.74, 0.07]}
\newcommand{\SsEExQwe}{$+$0.12}
\newcommand{\SsEExQweCI}{[$-$1.16, 1.67]}
\newcommand{\SsENGem}{26}
\newcommand{\SsENMis}{24}
\newcommand{\SsENQwe}{26}
\newcommand{\SsEOGem}{$+$2.13}

\newcommand{\SsEOMis}{$+$1.79}

\newcommand{\SsEOQwe}{$+$1.75}

\newcommand{\SsEXGem}{$+$2.20}

\newcommand{\SsEXMis}{$+$0.16}

\newcommand{\SsEXQwe}{$+$1.87}

\newcommand{\SsRExGem}{$+$1.18}
\newcommand{\SsRExGemCI}{[$-$0.02, 2.56]}
\newcommand{\SsRExMis}{$-$0.38}
\newcommand{\SsRExMisCI}{[$-$1.91, 1.20]}
\newcommand{\SsRExQwe}{$+$0.98}
\newcommand{\SsRExQweCI}{[$-$0.49, 2.62]}
\newcommand{\SsRNGem}{35}
\newcommand{\SsRNMis}{29}
\newcommand{\SsRNQwe}{30}

\newcommand{\TInGem}{$+$0.61}
\newcommand{\TInGemCI}{[0.00, 1.13]}
\newcommand{\TInMis}{$+$3.15}
\newcommand{\TInMisCI}{[2.71, 3.55]}
\newcommand{\TInQwe}{$+$0.39}
\newcommand{\TInQweCI}{[$-$1.50, 1.26]}
\newcommand{\TOutGem}{$-$0.45}
\newcommand{\TOutGemCI}{[$-$1.02, $-$0.02]}
\newcommand{\TOutMis}{$+$2.92}
\newcommand{\TOutMisCI}{[0.91, 3.54]}
\newcommand{\TOutQwe}{$-$0.42}
\newcommand{\TOutQweCI}{[$-$1.16, 0.00]}

\newcommand{\UNb}{330}
\newcommand{\UNn}{988}
\newcommand{\UNrate}{0.33}
\newcommand{\ViewAdj}{$+$0.09}
\newcommand{\ViewAdjCI}{[0.00, 0.17]}
\newcommand{\ViewRaw}{$+$0.13}

\title{Visible but Not Avoided: State-Dependent Re-Proposal of Rejected\\Candidates in Catalog-Based LLM Optimization Loops}
\author{Jianlin Chen \qquad Wenhui Chen \qquad Chi Man VONG}

\begin{document}
\maketitle
\begin{abstract}
LLM optimization loops often show the proposer the candidates it has already tried and rejected, but leave it to the proposer to avoid them. We ask whether avoidance measured in a static probe predicts avoidance in the states an optimizer's own trajectory produces. In catalog-based loops where three open LLMs improve a GUI agent by naming one of 40 edits, a re-proposal is identified exactly by the edit's name. Clean base-state probes underpredict re-proposal in the proposers' own loops (macro-averaged loop-minus-probe rate \DeltaMacro{} \DeltaMacroCI), and proposals of already evaluated harness--edit pairs consume \CostPctMin--\CostPctMax\% of valid proposer iterations. The two settings reject different edits: a probe's rejected edits are mostly ones the proposer did not choose, whereas a loop's are its own earlier choices, which remain preferred when the history is removed. Ordinary history lowers re-proposal for two proposers but raises it for the third, and the effects of failure evidence are model-specific. Stating the rejection status in one sentence lowers re-proposal in every model (\RfFN{} \RfFNCI{} relative to a neutral list of the same IDs). For catalog-based loops with exact edit identities, rejection avoidance should be measured on the rejected sets the loop produces.
\end{abstract}

\section{Introduction}\label{sec:intro}

\begin{figure*}[t]
\centering
\resizebox{\textwidth}{!}{
\begin{tikzpicture}[font=\footnotesize, >=Stealth, line width=0.6pt,
  comp/.style={draw=black!65, rounded corners=2pt, fill=black!4, minimum width=3.3cm, minimum height=1.0cm, align=center, inner sep=2pt},
  cool/.style={draw=blue!55!black, rounded corners=2pt, fill=blue!6, text width=6.6cm, minimum height=1.05cm, align=center, inner sep=2.5pt},
  warm/.style={draw=red!60!black, line width=1.1pt, rounded corners=2pt, fill=red!7, text width=6.6cm, minimum height=1.05cm, align=center, inner sep=2.5pt},
  link/.style={->, draw=black!55, densely dotted}]
\node[comp] (hm) at (0, 1.6) {\textbf{Reached harness + menu}\\[1pt]$h_s,\ M_s,\ X_s$};
\node[comp] (fh) at (0, 0)    {\textbf{Failure-bearing}\\\textbf{history}};
\node[comp] (rs) at (0,-1.6) {\textbf{Rejection signal}};
\node[draw=black!50, rounded corners=3pt, fit=(hm)(rs), inner sep=5pt,
      label={[font=\footnotesize\bfseries]above:Reached optimization state $s$}] (S) {};
\node[cool] (H) at (5.7, 1.6) {\textbf{No-history rejected-set preference} (H)\\[1pt]$X_s$ = the proposer's own earlier choices,\\still preferred without history\\[-1pt]{\scriptsize\color{black!70} no detectable excess over its other earlier choices}};
\node[cool, dashed] (F) at (5.7, 0) {\textbf{Failure evidence}: B, R, J, K, S\\[1pt]model-specific modulation};
\node[warm] (N) at (5.7,-1.6) {\textbf{Rejection signal}: N $\rightarrow$ F $\rightarrow$ P\\[1pt]explicit rejection status lowers\\re-proposal in every model\\[-1pt]{\scriptsize\color{black!70} added prohibition: further drop, n.s.\ after Holm}};
\draw[link] (hm.east) -- (H.west);
\draw[link] (fh.east) -- (F.west);
\draw[link] (rs.east) -- (N.west);
\node[draw=black!80, line width=0.9pt, rounded corners=2pt, fill=black!4, minimum width=2.3cm, minimum height=1.6cm, align=center] (P) at (10.9, 0)
     {\textbf{LLM proposer}\\[3pt]$p_s(e)$};
\draw[->, draw=blue!55!black] (H.east) .. controls +(0.6,0) and +(-0.6,0.4) .. ([yshift=5pt]P.west);
\draw[->, draw=blue!55!black, dashed] (F.east) -- (P.west);
\draw[->, draw=red!60!black, line width=1.1pt] (N.east) .. controls +(0.6,0) and +(-0.6,-0.4) .. ([yshift=-5pt]P.west);
\node[draw=black!60, rounded corners=2pt, fill=white, minimum width=2.3cm, minimum height=0.9cm, align=center] (E1) at (13.95, 0.8)
     {New edit\\$e\notin X_s$};
\node[draw=red!60!black, line width=1.4pt, rounded corners=2pt, fill=red!10, minimum width=2.3cm, minimum height=0.9cm, align=center] (E2) at (13.95,-0.8)
     {\textbf{Re-proposal}\\$e\in X_s$};
\draw[->, draw=black!60] (P.east) -- ++(0.35,0) |- (E1.west);
\draw[->, draw=red!60!black, line width=1.1pt] (P.east) -- ++(0.35,0) |- (E2.west);
\end{tikzpicture}}
\caption{\textbf{State-dependent re-proposal and the interventions used to localize it.} H: no history; B, R, J, K, S: failure-evidence edits; N, F, P: neutral ID list, status sentence, status plus prohibition. Without history, $X_s$ is strongly enriched, with no detectable excess over the proposer's other earlier choices (post hoc; \S\ref{sec:res-transport}). Failure-evidence edits have model-specific effects (\S\ref{sec:res-failure}). Stating rejection status lowers re-proposal in every model; a prohibition gives a further negative rate estimate, not significant after Holm correction ($p$ \RfPFP; \S\ref{sec:res-signal}). $\alpha=a-\lambda$ is a conceptual accounting; channel-specific readings require Assumption~\ref{ass:channel}.}
\label{fig:concept}
\end{figure*}
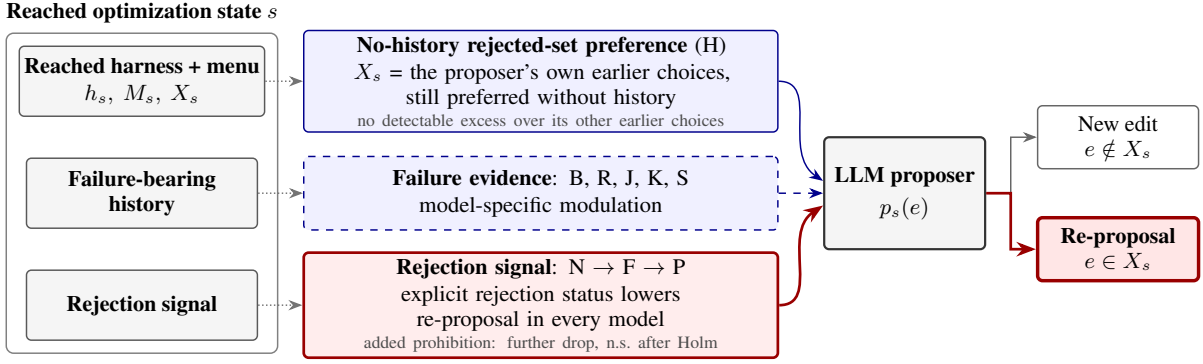

LLM optimizers improve prompts, programs and agent harnesses by proposing a change, evaluating it and recording the outcome for the next proposal~\citep{yang2024opro,khattab2023dspy,agrawal2026gepa,novikov2025alphaevolve,hu2024adas}. Many show the proposer its failed attempts and leave the avoidance to it: OPRO asks for solutions different from the shown ones~\citep{yang2024opro}, self-evolving agents keep rejected edits ``to discourage repetition''~\citep{lu2026procgraphs} or in a rejected-edit buffer~\citep{yang2026skillopt}, and Meta-Harness credits its gains to richer access to prior candidates~\citep{lee2026metaharness}. Showing a rejection is not enforcing it, as a tabu list does~\citep{glover1986future}. Under cached, deterministic evaluation, proposing an already evaluated candidate spends an iteration on a known outcome.

Avoidance of visible rejections is naturally tested on a fixed prompt, as instruction following is~\citep{zhou2023ifeval,jiang2024followbench}. Such a probe can show rejected candidates chosen without regard to the proposer's preferences. In a loop, the rejected candidates are the proposer's own earlier choices: it proposed them because they looked promising, and a failed evaluation need not change that judgement. A proposer can therefore pass over arbitrary rejected candidates in a probe and still return to its own rejected favourites in a loop. We call the loop behaviour \emph{visible-rejection avoidance}, reserving \emph{compliance} for an explicit prohibition.

We study this question where re-proposal is exact. A proposer improves a screenshot-only GUI agent by naming one of 40 catalog edits, so a re-proposal is identified exactly by the edit's name. We separate \emph{visible re-proposal}, choosing an edit shown as rejected with the current harness, from \emph{duplicate-pair proposal}, choosing any harness--edit pair already evaluated in the run. Three open proposers (Gemma-4, Mistral-Small-3.2, Qwen3-VL-32B) run the loop with two surface realizations of the catalog, which vary identifiers and presentation while holding the candidates and task fixed.

In a clean base-state probe, two proposers never chose a shown rejected edit and the third rarely did. In their own loops all three re-propose: the macro-averaged loop-minus-probe rate is \DeltaMacro{} \DeltaMacroCI, positive in all six model--catalog cells, and duplicate-pair proposals consume \CostPctMin--\CostPctMax\% of valid proposer iterations. Low re-proposal in a static probe does not guarantee low re-proposal in the states an optimizer reaches (\S\ref{sec:res-transport}).

To locate the gap, we rebuild sampled loop states. Keeping only the reached harness and the rejected entries comes close to the original state, passing the preregistered $\pm0.10$ equivalence margin but not $\pm0.05$. Removing the whole history block (scores, failures and rejection marks) shows that the rejected edits remain preferred: every proposer puts more mass on them than a uniform choice would. A post hoc comparison finds its other earlier choices enriched about as much, which points to self-selection without identifying a complete causal mechanism. Ordinary history lowers re-proposal for Gemma-4 and Qwen3-VL-32B and raises it for Mistral.

The history carries evidence about failures and the bare fact of rejection, so we edit the two separately on fresh states. Failure-evidence effects differ by model: consistent with failure content for Qwen3-VL-32B, mostly non-semantic for Mistral and small for Gemma-4. Rejection-status wording acts consistently: a sentence stating that the listed edits are in the harness or were tried with it and rejected lowers re-proposal relative to a neutral list of the same IDs in every model (\RfFN{} \RfFNCI). Adding ``Do not propose any of them'' lowers the rate further (\RfPF), but this step did not pass the registered multiplicity-corrected test (Holm $p$ \RfPFP; \S\ref{sec:res-failure}--\ref{sec:res-signal}). Reading these wording effects as changes in an internal exclusion process requires the assumptions of \S\ref{sec:model}.

\paragraph{Contributions.}
\begin{enumerate}[leftmargin=*,itemsep=1pt,topsep=2pt]
\item A measurement framework that separates visible re-proposal from duplicate-pair proposal, scores re-proposal against a uniform choice over the menu, and states what probe-to-loop comparisons and matched-state edits identify (\S\ref{sec:model}).
\item Evidence across three proposers that clean-probe avoidance does not transport to the rejected sets their own loops produce, and that these rejected edits remain preferred without history (\S\ref{sec:res-transport}).
\item Matched-state interventions showing model-specific failure-evidence effects and a consistent effect of rejection-status wording (\S\ref{sec:res-failure}--\ref{sec:res-signal}).
\end{enumerate}

\section{Related Work}\label{sec:related}

\paragraph{LLM optimizers and the shown-history premise.} Prompt optimizers propose instructions with an LLM and select them by score~\citep{zhou2023ape,pryzant2023protegi,guo2024evoprompt,fernando2023promptbreeder,khattab2023dspy,opsahlong2024mipro,yuksekgonul2024textgrad,yang2024opro,agrawal2026gepa}. Program and agent optimizers keep archives of evaluated candidates~\citep{romeraparedes2024funsearch,novikov2025alphaevolve,lange2025shinka,lee2025mindevolution,hu2024adas,zhang2025dgm,lee2026metaharness,xia2026rrsi}. Whether these systems work, and why, has been examined through their scores and trajectories~\citep{ma2024promptopt,zhang2024revisitopro,zhang2026goodoptimizer}. Classical search treats exclusion as part of the algorithm, through tabu lists and novelty archives~\citep{glover1986future,lehman2011novelty}. Some LLM systems enforce it as well, for example by rejection sampling on code novelty~\citep{lange2025shinka}, but most show the history and expect the proposer to act on it. How a system should define and enforce novelty is studied in a related manuscript~\citep{anon2027cin}; here the proposer is only shown its history. We test that expectation at the states these loops reach. \citet{zhang2026goodoptimizer} find that zero-shot ability explains only part of an optimizer's trajectory. We isolate one transport failure, rejection avoidance, with exact candidate identity and matched-state counterfactuals.

\paragraph{Instruction following and negative constraints.} Instruction-following benchmarks score one response against constraints written into one prompt~\citep{zhou2023ifeval,jiang2024followbench,qin2024infobench,zhang2026firebench}. Negation is a known weakness~\citep{jang2022negated,garciaferrero2023negation}, telling a model to avoid an entity can elicit it~\citep{castricato2024pink}, prohibitions decay over long conversations~\citep{gamage2026omission}, and context compaction can erase constraints~\citep{chen2026governance}. \citet{rana2026gravity} explain violations of a word prohibition by the model's prior pressure towards the word. Our no-history baseline measures such pressure at the states a loop reaches, and shows that it is directed at the proposer's own earlier choices.

\paragraph{Failure records and repetition in context.} Language models repeat what is in their context: induction heads copy earlier tokens~\citep{olsson2022induction}, decoding that maximises likelihood produces repetitive text~\citep{holtzman2020degeneration}, and a repeated sentence becomes more likely the more often it appears~\citep{xu2022repetition}. For agents, a failed attempt left in context raises the error rate of a retry~\citep{yang2026retrying}. The closest work is that of \citet{gumaan2026backfires}. In 135M--1.7B tool-using agents, a failure record raises the probability of repeating the failed call over a fixed candidate set, mostly through the surface form of the call, and a ``do not repeat'' instruction leaves the effect unchanged. Our primary distinction is the object of study: an evolving, trajectory-generated rejected set in an optimization loop, measured by exact identity at the states the loop actually reaches. Unlike their ``do not repeat'' instruction, our DO-NOT sentence lowers re-proposal substantially, and most of the reduction is already achieved by stating the rejection status. Reflection and stored experience help agents when feedback is informative~\citep{shinn2023reflexion,madaan2023selfrefine,zhao2024expel,alazraki2025mistakes,huang2024selfcorrect,kamoi2024selfcorrect}, while irrelevant context distracts~\citep{shi2023distracted}.

\paragraph{Knowing versus acting, reasoning, and long context.} Models recall the constraints they violate~\citep{kruthof2026drift}, show a gap between accessible memory and action~\citep{dong2026memarbiter}, and act unfaithfully to their own execution history~\citep{zhang2025mirage}. Our proposers show a related dissociation: they justify re-proposals by the failures they see, yet for two of three models the content of those failures has little measurable effect on the choice (\S\ref{sec:res-failure}). Stronger reasoning lowers compliance with static instructions~\citep{fu2025mathif,li2025thinkingfails}, and reasoning traces rarely follow instructions about themselves~\citep{kwon2026reasonif}. Long histories also suffer from weak use of the middle of the context, short effective context, position effects and failed in-context exploration~\citep{liu2024lost,hsieh2024ruler,guo2024serial,krishnamurthy2024explore}. App.~\ref{app:matrix} lists each line of work with the assumption we test.

\section{Measurement and Behavioural Accounting Framework}\label{sec:model}

This framework was developed after the first experiments, whose registered endpoints were rate differences (App.~\ref{app:prereg}); the fresh-state round (\S\ref{sec:design-failure}--\ref{sec:design-signal}) was registered afterwards.

\subsection{States and repeated proposals}\label{sec:model-state}

A catalog loop keeps a harness $h_t\subseteq E$ and asks the proposer for one edit from the menu $M_t=E\setminus h_t$. The edit is kept if development accuracy rises by more than $\delta$; either way the attempt, its outcome and some failures are appended to the history the proposer sees, so a rejected edit can be proposed again.

\begin{definition}[State and visible rejection]\label{def:state}
The state is $s_t=(h_t,H_t,q_t)$, with history $H_t$ and rendered prompt $q_t$. The visible rejected set $X_t\subseteq M_t$ contains the edits shown in $q_t$ as tried with harness $h_t$ and not kept. The proposer's choice follows the serving distribution $p_s(e)=P_{\mathrm{serve}}(e_t=e\mid s,\text{endpoint},\text{decoding},\text{valid reply})$ over $M_s$, which includes provider and serving nondeterminism. A re-proposal is $B_t=\mathbf{1}[e_t\in X_t]$, with rate $\beta(s)=p_s(X_s)$; a state is eligible if $\varnothing\neq X_t\neq M_t$. We write $s$ for a generic state and $X_s,M_s,h_s$ for its components.
\end{definition}

Replies naming no catalog edit are invalid and excluded (harness edits: App.~\ref{app:protocol}). A visible re-proposal ($B_t=1$) chooses an edit in the visible rejected set. A duplicate-pair proposal, $D_t=\mathbf{1}[(h_t,e_t)$ evaluated earlier in the run$]$, chooses an evaluated pair whether or not its rejection is still visible: $B_t=1$ implies $D_t=1$, but under the truncated W8 view $D_t=1$ can occur with $B_t=0$. Under cached evaluation, $D_t$ counts iterations spent without exploring a new pair.

\subsection{A reference for re-proposal}\label{sec:model-alpha}

States differ in how much of the menu is rejected, so a higher raw rate need not mean a stronger relative preference. As an illustration, not data: if 8 of 40 available edits are rejected, a uniform choice re-proposes with probability 0.20, and a proposer that puts 0.35 on the rejected set exceeds this reference.

\begin{definition}[Rejection-enrichment log-odds]\label{def:alpha}
For a state $s$ and a target set $T\subseteq M_s$ with $0<|T|<|M_s|$, the enrichment log-odds of $T$ are $\alpha_T(s)=\operatorname{logit}p_s(T)-\operatorname{logit}(|T|/|M_s|)$, where $|T|/|M_s|$ is the probability that a uniform choice over the menu lands in $T$. For the visible rejected set, the uniform-menu reference is $\beta^{U}(s)=|X_s|/|M_s|$, and the rejection-enrichment log-odds are $\alpha(s)\equiv\alpha_{X_s}(s)=\operatorname{logit}\beta(s)-\operatorname{logit}\beta^{U}(s)$.
\end{definition}

Eligibility gives $0<|X_s|<|M_s|$, so $\alpha$ is always defined. $\alpha>0$ means more mass on rejected edits than the uniform reference, $\alpha<0$ less and $\alpha=0$ the same (the illustration has $\alpha\approx0.77$); $\alpha=\pm\infty$ when $\beta(s)\in\{0,1\}$. The reference removes the rejected fraction of the menu, not other differences between states.

\subsection{Preference without history}\label{sec:model-nohist}

Do a reached state's rejected edits remain attractive without the history? Let $s^{\varnothing}$ have the same harness, menu and question but an empty history. The whole history block (scores, failures, rejection marks, accuracy trajectory) is removed, so $s^{\varnothing}$ does not isolate rejection information. Its own rejected set is empty; we evaluate the source state's $X_s$ as a target set, $\alpha_{X_s}(s^{\varnothing})=\operatorname{logit}p_{s^{\varnothing}}(X_s)-\operatorname{logit}\beta^{U}(s)$. Because the rejected set contains the proposer's earlier choices at the current harness, stable preferences can yield positive enrichment even when history is removed; we compare it with the proposer's other earlier choices (\S\ref{sec:res-transport}). It measures preference over a self-selected set, not a causal effect or a specific mechanism.

\subsection{Accounting for changes between states}\label{sec:model-account}

\begin{assumption}[Exclusion-penalised choice]\label{ass:choice}
For every state considered in the accounting, including matched counterfactual states, $p_s(e)\propto w_s(e)\,\exp\bigl(-\lambda_s\,\mathbf{1}[e\in X_s]\bigr)$ for $e\in M_s$, with attraction weights $w_s(e)\ge0$ and an exclusion strength $\lambda_s\ge0$.
\end{assumption}

\emph{Attraction} is shorthand for residual relative choice weight, not a cognitive mechanism: $w_s(e)$ absorbs every factor other than the explicit penalty. Single weights may be zero, but an evaluated target set $T$ and $M_s\setminus T$ need positive total weight. Any distribution with $0<\beta(s)<1$ has this form with $\lambda_s=0$ and $w_s=p_s$, so the assumption is a parametrisation (weights up to a common factor): $a$ and $\lambda$ are not identified from choices, and Identities~\ref{prop:decomp}--\ref{prop:probe} follow algebraically from Definition~\ref{def:alpha}.

\begin{identity}[Decomposition]\label{prop:decomp}
Under Assumption~\ref{ass:choice}, $\alpha(s)=a(s)-\lambda_s$, where $a(s)\equiv a_{X_s}(s)$ and $a_T(s)=\log\bar w_s(T)-\log\bar w_s(M_s\setminus T)$ is the log ratio of the mean attraction inside and outside a set $T$.
\end{identity}

\begin{identity}[Probe transport]\label{prop:probe}
For eligible states $s_0$ and $s$, write $\Delta f=f(s)-f(s_0)$. Under Assumption~\ref{ass:choice},
\[\Delta\operatorname{logit}\beta=\Delta a-\Delta\lambda+\Delta\operatorname{logit}\beta^{U},\]
and the corresponding enrichment identity is $\Delta\alpha=\Delta a-\Delta\lambda$.
\end{identity}

Re-proposal thus differs between states through the rejected fraction of the menu ($\Delta\operatorname{logit}\beta^{U}$), which the reference removes, and through enrichment, written as a change in attraction ($\Delta a$) minus a change in exclusion ($\Delta\lambda$). For a fixed reference, $\beta^{U}(s)=\beta^{U}(s_0)$, a probe at $s_0$ bounds re-proposal at $s$ exactly when attraction has not risen by more than exclusion. In $s^{\varnothing}$ no edit is marked as rejected, so $\alpha_{X_s}(s^{\varnothing})=a_{X_s}(s^{\varnothing})$, and the history contrast
\[\alpha_{X_s}(s)-\alpha_{X_s}(s^{\varnothing})=[a_{X_s}(s)-a_{X_s}(s^{\varnothing})]-\lambda_s\]
is the net effect of the ordinary history on the mass assigned to $X_s$; it does not separate attraction from exclusion.

\subsection{What matched-state edits identify}\label{sec:model-ident}

A matched-state edit's effect on re-proposal is observed directly; attributing it to attraction or exclusion needs a further assumption.

\begin{assumption}[Channel separation]\label{ass:channel}
A state edit $\phi$ leaves $h_s$, $X_s$ and $M_s$ unchanged. An evidence edit (changing the failures shown with history entries) leaves $\lambda_s$ unchanged. Two sentences listing the same IDs at the same position change $a(s)$ equally.
\end{assumption}

\begin{proposition}[Identification under channel separation]\label{prop:ident}
Under Assumptions~\ref{ass:choice}--\ref{ass:channel}, the log odds ratio of re-proposal between $\phi(s)$ and $s$ equals $\alpha(\phi(s))-\alpha(s)$; for an evidence edit it equals $a(\phi(s))-a(s)$, and between two sentence edits $\phi_1,\phi_2$ listing the same IDs at the same position (for example F and N, or P and F) it equals $-(\lambda_{\phi_1(s)}-\lambda_{\phi_2(s)})$.
\end{proposition}

Proofs are in App.~\ref{app:proofs}. Assumption~\ref{ass:channel} is not identified from choices, so the contrasts identify effects on observed re-proposal and their attribution to a channel is conditional on it. Matched controls test its implications: if deleting failure spans acts through their content, an approximately length-matched neutral replacement should behave like their absence (\S\ref{sec:res-failure}).

\paragraph{Estimation.}\label{sec:model-est} $\alpha$ is estimated by logistic regression of $B_t$ with offset $\operatorname{logit}\beta^{U}(s_t)$, or by an exact one-sided 95\% upper bound when no re-proposal occurs. Matched-state contrasts are rate differences, also reported as Mantel--Haenszel log odds ratios stratified by state (the sample analogue of Proposition~\ref{prop:ident}), with state-level and hierarchical bootstrap CIs (App.~\ref{app:proofs}).

\section{Experimental Design}\label{sec:setup}

\subsection{Optimization setting}\label{sec:design-setting}

The loop improves a screenshot-only GUI agent, Qwen3-VL-8B-Instruct~\citep{bai2025qwen3vl}, on phone screens from nine synthetic apps with numbered set-of-mark tags~\citep{yang2023som}; the agent replies \texttt{TAP} $n$ or \texttt{ANSWER} \emph{text}. A harness is a set of edits from a 40-edit catalog (29 prompt, 9 rendering and 2 decoding edits), scored on 60 development items; agent calls are cached, so each harness has one accuracy (base: 0.75). For 30 iterations the proposer sees the task, the catalog, the current harness with its accuracy and the history, and names one edit, which is kept if accuracy rises by more than $\delta=0.02$. Each history entry gives the edit, the new accuracy, \emph{kept} or \emph{not kept}, and two failed items (instruction, agent reply, correct tag). The view is the last 8 entries (W8) or the full history (FULL); decoding uses temperature 0 and at most 200 tokens. The catalog has two independently written surface realizations of the same 40 edits: catalog~A with IDs \texttt{E01}--\texttt{E40} and terse lines, catalog~B with codes such as \texttt{XR4}, reworded and grouped descriptions and different task, question and history wording. The two versions hold candidate semantics and domain fixed while varying identifiers and presentation~\citep{sclar2024formatting} (App.~\ref{app:catalogs}). Gemma-4-26B-A4B-it, Mistral-Small-3.2-24B-Instruct and Qwen3-VL-32B-Instruct (text-only prompts), accessed through OpenRouter, run 36 loops: 3 models $\times$ 2 catalogs $\times$ 2 views $\times$ 3 seeds.

\subsection{Does static avoidance transport to reached states?}\label{sec:design-transport}

\emph{Clean probe.} At the base harness, the probe finds the model's zero-history favourite (the first rejected edit of its removal chain) and shows a window of $L\in\{8,16\}$ real rejected base evaluations with their failures, the favourite at recency position $d\in\{1,L/2,L\}$ and the other entries drawn at random; five catalog orders give 30 cells per model and catalog. The clean probe and the reached loop states differ in harness, history and rejected-set composition. \emph{Own loops.} We record $B_t$ at every eligible step of the 36 loops. \emph{Reached-state reconstruction.} From the eligible loop steps we sample states uniformly, regardless of the reply: 120 states in a first round and 360 fresh states in a second (\S\ref{sec:design-registration}). Condition M keeps the harness and the rejected entries with their failures and drops every other entry. Condition H keeps the harness, catalog and question and empties the history (\S\ref{sec:model-nohist}); its replies are scored against the original $X_s$.

\subsection{Edits of the failure evidence}\label{sec:design-failure}

On the same sampled states (overview: Fig.~\ref{fig:concept}), with the unchanged state (A) as reference: \emph{presence}, B deletes the failures from every entry; \emph{text bulk}, R replaces each failure segment with approximately length-matched neutral text (a rural passage with no digits or task words) inserted at the same position with the same entry layout, within 13 characters in the realized prompts; \emph{type}, C replaces failures with unshown failures of the same evaluation and type, D with failures without an ordinal or positional word (``eighth'', ``row''), and T, on the 13 states where one exists, with a non-ordinal failure of the same evaluation, app and length ($\pm10$ characters); \emph{attachment}, J deletes failures only from the entries of $X_s$, K only from the other entries, and S permutes the failure pairs among the entries of $X_s$.

\subsection{Edits of the rejection signal}\label{sec:design-signal}

Each condition adds one sentence before the question, listing the IDs that are in the harness or rejected with it: N a neutral list (``For reference, the following edit IDs appear in this prompt: \ldots''); F a status sentence (``The following edits are already in the current harness or were already tried with it and rejected: \ldots. This note records their status.''); P the same status sentence ending ``Do not propose any of them.''; and E the original sentence (``DO NOT PROPOSE any of the following edits. They are already in the current harness or were already tried with it and rejected: \ldots''). The contrasts F $-$ N and P $-$ F separately test the incremental effects of rejection-status wording and of an explicit prohibition, with the listed IDs and prompt position held fixed; mapping these effects to the exclusion channel remains conditional on Assumption~\ref{ass:channel}.

\subsection{Boundary and supporting analyses}\label{sec:design-boundary}

The regime comparison answers 40 fresh states with Qwen3-30B-A3B-Instruct-2507 and Qwen3-30B-A3B-Thinking-2507~\citep{yang2025qwen3} under the same 4000-token cap, first with unpinned routing and then with both endpoints pinned to one provider and three seeds. For stated reasons, two blinded LLM annotators coded 60 replies with a registered eight-code taxonomy (App.~\ref{app:codebook}).

\subsection{Registration and inference}\label{sec:design-registration}

Every experiment was preregistered in a locally timestamped JSON file before its first paid call (App.~\ref{app:prereg}). The loops, probes and the first 120 states (A--E) were registered before any analysis, while M and N on those states are follow-up tests registered after A--E were observed. The 360 fresh states (A, H, M, B, R, S, J, K, T, N, F, P, E on 120; A and M on 240 more) were registered after the framework of \S\ref{sec:model}, with one-sided hypotheses for B, R, J $-$ K, S, P $-$ F and E (Holm over six), two-sided ones for A $-$ H, F $-$ N and R $-$ B (Holm over three), and an equivalence test of M $-$ A at $\pm0.10$ (secondary $\pm0.05$). Per-model contrasts were registered as descriptive, without multiplicity correction, and the self-selection comparison of \S\ref{sec:res-transport} is post hoc. Each state and condition is queried three times at temperature 0 with provider seeds 1--3, and the unit is the state. Loops are resampled by run, probes by order, and states by state and by run, and tests are sign-flip tests. All paid calls cost \$\SpentTotal.

\section{Results}\label{sec:results}

\subsection{Static avoidance does not transport to the rejected sets loops produce}\label{sec:res-transport}

\begin{table*}[t]
\centering\small\setlength{\tabcolsep}{4pt}
\begin{tabular}{@{}llrlrll@{}}
\toprule
 & & \multicolumn{2}{c}{Own loops} & \multicolumn{2}{c}{Clean probe} & \\
\cmidrule(lr){3-4}\cmidrule(lr){5-6}
Proposer & Cat. & $n_B/n$ & $\hat\alpha$ [95\% CI] & $n_B/n$ & $\hat\alpha$ & $\Delta$ [95\% CI] \\
\midrule
Gemma-4-26B-A4B & A & 22/130 & $+$0.43 [$-$0.56, 1.21] & 0/30 & $\le$$-$1.46 & 0.17 [0.08, 0.26] \\
Gemma-4-26B-A4B & B & 34/156 & $+$0.30 [$-$0.27, 0.84] & 0/30 & $\le$$-$1.46 & 0.22 [0.13, 0.31] \\
Mistral-Small-3.2-24B & A & 121/167 & $+$3.52 [3.27, 3.69] & 2/30 & $-$1.84 & 0.66 [0.52, 0.75] \\
Mistral-Small-3.2-24B & B & 94/158 & $+$2.69 [1.98, 3.38] & 3/30 & $-$1.40 & 0.49 [0.31, 0.66] \\
Qwen3-VL-32B & A & 16/161 & $-$0.89 [$-$1.74, $-$0.30] & 0/30 & $\le$$-$1.46 & 0.10 [0.05, 0.16] \\
Qwen3-VL-32B & B & 42/152 & $+$0.61 [$-$0.26, 1.23] & 0/30 & $\le$$-$1.46 & 0.28 [0.13, 0.42] \\
\midrule
Macro (6 cells) & & 329/924 & & 5/180 & & 0.32 [0.27, 0.37] \\
\bottomrule
\end{tabular}

\caption{\textbf{Probe and loop.} $n_B/n$: re-proposals over eligible valid loop steps (6 runs per cell) or valid probe cells. $\hat\alpha$: rejection-enrichment log-odds with run-cluster CI; $\le$: exact one-sided 95\% upper bound for a probe with no re-proposal. $\Delta$: registered primary endpoint, loop rate minus probe rate, two-level cluster bootstrap CI.}
\label{tab:p01}
\end{table*}

\begin{figure}[t]
\centering
\includegraphics[width=\columnwidth]{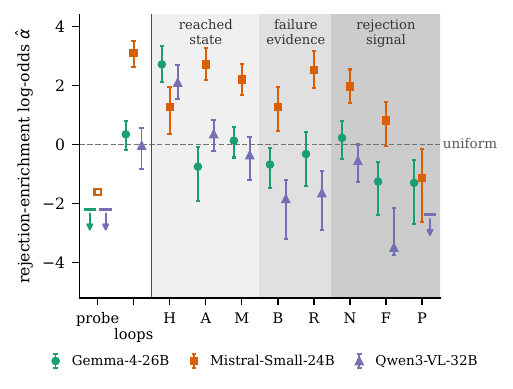}
\caption{\textbf{Rejection-enrichment log-odds} against a uniform choice (dashed); arrows: exact 95\% upper bounds. H--P: conditions on the 120 fresh states (Table~\ref{tab:p02}); state-cluster CIs.}
\label{fig:avoid}
\end{figure}

In the clean probe all three proposers avoid the window: Gemma-4 and Qwen3-VL-32B named no window edit in any of their 60 cells ($\alpha\le$\APubGem), and Mistral named one in \MisProbeB{} of \MisProbeN{}. In their own loops they re-propose: the registered loop-minus-probe difference is \DeltaMacro{} \DeltaMacroCI, positive in \HoneB{} of 6 cells (Table~\ref{tab:p01}), and robust to how invalid replies are counted (\DeltaInvZero{} to \DeltaInvOne). Against the uniform-menu reference, Gemma-4 and Qwen3-VL-32B choose rejected edits about as often as a uniform choice would (\ALGem, \ALQwe) and Mistral far more often (\ALMis; Fig.~\ref{fig:avoid}). Proposals of already evaluated harness--edit pairs ($D_t=1$) consume \CostPctGem\% (\CostRedGem/\CostValGem), \CostPctQwe\% (\CostRedQwe/\CostValQwe) and \CostPctMis\% (\CostRedMis/\CostValMis) of valid proposer iterations (App.~\ref{app:verify}).

The probe and the loop differ in harness, rejected set and history. The same-harness reconstruction (M) keeps the first two and drops other history. On 360 fresh states its rate is \RfEqM, against \RfEqA{} for the unchanged state (A). M $-$ A is \RfEq{} (90\% CI \RfEqCI; hierarchical \RfEqCIh), meeting the preregistered $\pm0.10$ equivalence margin but not $\pm0.05$. The A and M prompts are identical in \RfSameN{} states. In the \RfChangedN{} states where M removes history, a post hoc analysis gives M $-$ A \RfChanged{} (90\% CI \RfChangedCI; hierarchical \RfChangedCIh; App.~\ref{app:supp}). The hierarchical interval crosses the $\pm0.10$ boundary, so equivalence is established for the registered average over all states, not for every state where the prompt changes. The base probe remains \DeltaMacro{} below the loops.

With history removed (H), every proposer assigns more probability mass than uniform to the source-state rejected set $X_s$: \RaHGem{} (Gemma-4), \RaHMis{} (Mistral) and \RaHQwe{} (Qwen3-VL-32B). This is consistent with self-selection, since $X_s$ consists of the proposer's own earlier choices at $h_s$. In a post hoc comparison on the states where it had earlier chosen edits at other harnesses, those edits are enriched about as much without history (\SsEOGem, \SsEOMis, \SsEOQwe), and $X_s$ shows no detectable excess over them (\SsEExGem{} \SsEExGemCI, \SsEExMis{} \SsEExMisCI, \SsEExQwe{} \SsEExQweCI; App.~\ref{app:verify}). The rejected set thus consists of the proposer's earlier choices, which remain preferred relative to the uniform-menu reference when history is removed. Ordinary history lowers re-proposal relative to no history for Gemma-4 (\RfAHGem{} in rate) and Qwen3-VL-32B (\RfAHQwe), though not to probe levels, and raises it for Mistral (\RpAHMis{} \RpAHMisCI).

The probe's window holds one preferred edit, which the proposer avoids by choosing another, and \PCSGem{} of \PCSGemN, \PCSMis{} of \PCSMisN{} and \PCSQwe{} of \PCSQweN{} loop re-proposals name one of a model's three most frequent probe choices. In the loops, avoidance fails when a model's two most frequent probe choices are themselves in $X_t$ (post hoc; $\hat\alpha$ without and with them: Gemma-4 \TOutGem{} and \TInGem, Qwen3-VL-32B \TOutQwe{} and \TInQwe; Mistral is attracted either way). These observations suggest that rejected-set composition contributes to the probe--loop gap, although the post hoc comparisons do not quantify its causal contribution.

\subsection{Failure evidence has model-specific text-bulk and content effects}\label{sec:res-failure}

\begin{table*}[t]
\centering\small\setlength{\tabcolsep}{3.5pt}
\resizebox{\textwidth}{!}{\begin{tabular}{@{}llrlrrlr@{}}
\toprule
 & & \multicolumn{3}{c}{Original 120 states} & \multicolumn{3}{c}{Fresh R4 intervention subset (120 states)} \\
\cmidrule(lr){3-5}\cmidrule(lr){6-8}
Role & Condition & Rate & $X-A$ [95\% CI] & log OR & Rate & $X-A$ [95\% CI] & log OR \\
\midrule
Reference & A: unchanged state & 0.34 & -- & -- & 0.30 & -- & -- \\
\midrule
Baseline preference & H: no history (matched) & & & & 0.51 & $+$0.21 [0.11, 0.31] & $+$0.68 \\
\midrule
Localization & M: same-harness probe & 0.30 & $-$0.05 [$-$0.11, 0.01] & $-$0.35 & 0.27 & $\Delta_{360}$ $-$0.04 [$-$0.07, $-$0.01]$^\dagger$ & $-$0.36$^\dagger$ \\
\midrule
Failure evidence & B: failure examples removed & 0.15 & $-$0.19 [$-$0.25, $-$0.13] & $-$1.92 & 0.14 & $-$0.17 [$-$0.23, $-$0.10] & $-$1.49 \\
 & R: neutral text, approx.\ length-matched & & & & 0.24 & $-$0.07 [$-$0.13, $-$0.01] & $-$0.56 \\
 & J: failures removed from $X_t$ entries only & & & & 0.25 & $-$0.05 [$-$0.12, 0.01] & $-$0.50 \\
 & K: failures removed from other entries only & & & & 0.30 & $-$0.00 [$-$0.04, 0.04] & $-$0.09 \\
 & S: failures shuffled across $X_t$ entries & & & & 0.32 & \multicolumn{2}{l}{follow 0.03$\to$0.05} \\
 & C: same-type failures & 0.37 & $+$0.04 [$-$0.03, 0.10] & $+$0.29 & & & \\
 & D: other-type failures & 0.22 & $-$0.11 [$-$0.18, $-$0.05] & $-$1.00 & & & \\
 & T: strict other-type swap ($n=13$) & & & & 0.54 & $-$0.10 [$-$0.33, 0.10] & $-$0.47 \\
\midrule
Rejection signal & N: neutral ID list & 0.27 & $-$0.06 [$-$0.13, 0.01] & $-$0.45 & 0.24 & $-$0.06 [$-$0.13, 0.00] & $-$0.50 \\
 & F: status sentence & & & & 0.09 & $-$0.22 [$-$0.28, $-$0.16] & $-$2.38 \\
 & P: status sentence $+$ prohibition & & & & 0.03 & $-$0.27 [$-$0.33, $-$0.21] & $-$2.88 \\
 & E: original DO-NOT sentence & 0.04 & $-$0.30 [$-$0.36, $-$0.24] & $-$3.20 & 0.03 & $-$0.27 [$-$0.32, $-$0.22] & $-$3.61 \\
\bottomrule
\end{tabular}
}
\caption{\textbf{State edits, by role.} Rate: mean of state-level rates (3 calls per state). $X-A$: state-level difference with stratified bootstrap CI ($^\dagger$: 90\% CI over 360 states for the equivalence test). log OR: Mantel--Haenszel, stratified by state. S: share of choices that follow a shuffled failure pair. For M, the displayed rate is from the 120-state subset; the daggered equivalence contrast and log OR ($\Delta_{360}$) use all 360 fresh states, on which the unchanged and reconstructed rates are \RfEqA{} and \RfEqM. Subset-defined contrasts (T, on the $n$ states where a strict swap exists; J $-$ K in App.~\ref{app:prereg}) are paired against A on the same subset; the overall A rate cannot be used to reconstruct them. Registered tests and hierarchical CIs: App.~\ref{app:prereg}.}
\label{tab:p02}
\end{table*}

Deleting the failures from every entry lowers re-proposal, on the first states (\Htwoa{} \HtwoaCI) and on the fresh ones (\RfBa{} \RfBaCI; Table~\ref{tab:p02}). The approximately length-matched control weakens a common semantic-content explanation of the deletion effect. Replacing the deleted spans with neutral text restores much of the pooled re-proposal rate, but the pattern differs sharply across models (R $-$ A \RfRa{} \RfRaCI, Holm $p$ \RfRaP; R $-$ B \RfRB{} \RfRBCI). Qwen3-VL-32B retains a substantial effect under neutral replacement, consistent with a content-sensitive component (R $-$ A \RfRaQwe). For Mistral, most of the deletion effect is recovered by restoring a neutral span (\RfBaMis{} for deletion, \RfRaMis{} for neutral text), suggesting a largely non-semantic text-bulk or span-presence effect. Gemma-4 shows little effect from either (\RfBaGem, \RfRaGem). The neutral replacement thus shows that deletion does not isolate semantic failure content. Consequently, B $-$ A should not be interpreted as a pure attraction contrast.

We find no clear evidence that re-proposals systematically follow the history entries to which failure examples are attached. Deleting failures only from the rejected entries (J) or only from the others (K) has small effects (\RfJa{} and \RfKa), with the registered contrast J $-$ K in the wrong direction (\RfJK{} \RfJKCI), and after a shuffle of failures across rejected entries, choices follow the moved failures no more often than before (\RfFolA{} vs.\ \RfFolS; Holm $p$ \RfFolP). Other-type replacements lowered re-proposal in the first round (D $-$ A \Htwod), but type is nearly confounded with app, and a strict within-app swap, possible on only \RfWithT{} fresh states, is inconclusive (\RfTa{} \RfTaCI). Failure-bearing history matters, but the evidence that its content drives re-proposal is limited to one model.

\subsection{Restating rejection status lowers re-proposal across all three main proposers}\label{sec:res-signal}

The no-history results show that previously selected edits can remain preferred. We therefore test whether stating their status changes proposals, with the listed IDs and their position held fixed. Repeating the IDs without saying what they are (N) changes little (\RfNA{} \RfNACI). Saying that they are in the harness or were tried with it and rejected (F) lowers re-proposal from \RrA{} to \RrF. Against the neutral list, F $-$ N is \RfFN{} \RfFNCI{} (log odds ratio \RfFNLo), and it is negative in every model (\RfFNGem, \RfFNMis, \RfFNQwe). Adding ``Do not propose any of them'' (P) lowers it further to \RrP. In rate this step is small and the registered test is not supported (P $-$ F \RfPF{} \RfPFCI, Holm $p$ \RfPFP), because little re-proposal remains after F, and it is carried by Mistral (\RfPFMis; Gemma-4 \RfPFGem, Qwen3-VL-32B \RfPFQwe). In log odds it is substantial (\RfPFLo{} \RfPFLoCI), comparable to the status step (\RfFNLo). The original DO-NOT sentence behaves like P (\RfEa{}; \Htwob{} on the first states).

F brings rejected edits below the uniform reference for Gemma-4 and Qwen3-VL-32B ($\hat\alpha$ \RaFGem, \RaFQwe), not for Mistral (\RaFMis). Reading F $-$ N as stronger exclusion requires Assumption~\ref{ass:channel}.

\subsection{Boundary conditions and supporting evidence}\label{sec:res-boundary}

\paragraph{Proposer regime.} On 40 fresh states with the same 4000-token cap, the thinking endpoint re-proposed in \ArmTfourB{} of \ArmTfourV{} states and the instruct endpoint in \ArmIfourB{} of \ArmIfourV{} (\Hfoura{} \HfouraCI; App.~\ref{app:supp}); a 200-token instruct cap changed nothing detectable (\Hfourb{} \HfourbCI). Pinned to \RgProv{} with three seeds, the rates were \RgT{} and \RgI{} (\Rg{} \RgCI), so the difference is not a routing artefact; the endpoints also differ in post-training and completion tokens (\RgTokT{} vs.\ \RgTokI).

\paragraph{Stated reasons.} Among sampled re-proposals, the most frequent code is the preregistered rationale code \emph{failure-target attraction}, a response that cites a failed item and proposes an edit to address it (\FTAvBD{} and \FTAvBG{} of 30 for the blinded pair, $\kappa=$\KappaBlind), whereas explicit retry is rare (\RTvBD{} and \RTvBG{} of 30). The label describes the stated rationale, not the attachment-level causal mechanism tested and not supported in \S\ref{sec:res-failure}.

\section{Discussion}\label{sec:discussion}

A static probe asks whether a proposer avoids rejected candidates it mostly had little reason to choose; a loop asks it to avoid its own rejected favourites. The same-harness reconstruction meets the registered equivalence margin on average, while the post hoc analysis of changed prompts shows larger effects for Mistral and Qwen3-VL-32B (App.~\ref{app:supp}). The comparison does not separate the effects of the harness and rejected-set composition. The self-selection comparison is post hoc and rests on few states, and history moves the three proposers in different directions, so no single mechanism describes them all. For catalog-based loops that delegate avoidance to the proposer, the results support evaluating it on rejected sets of the composition the loop produces and stating rejection status explicitly; systems needing hard non-repetition may prefer deterministic filtering, which we do not evaluate.

\section{Conclusion}\label{sec:conclusion}

A rejection that an LLM proposer can see is not necessarily one it avoids. In our catalog-based loops, proposers without history put more probability on a loop's rejected set than a uniform choice would; our clean probes only partially represent such self-selected sets, and stating rejection status lowers re-proposal in every model. For loops with exact candidate identities, avoidance should be measured on the rejected sets those loops produce.

\section*{Limitations}

\paragraph{Scope.} All loops improve one synthetic GUI agent with one 40-edit candidate set. Catalog~B is a second surface realization of the same candidates, so the results show robustness to presentation, not generalisation across candidate sets or domains. Free-text optimizers, where re-proposals cannot be matched exactly by name, are not studied.

\paragraph{Identification.} The framework was developed after the first experiments, and the fresh round tests it but uses the same loops. The no-history baseline assumes that an empty history leaves the proposer's preference over the menu as it would be at that harness, and it also removes the accuracy trajectory. Because $X_s$ consists of the proposer's own earlier choices, the no-history condition measures preference over a self-selected set. The comparison with the proposer's other earlier choices is post hoc and covers only the states where such choices exist (\SsENGem, \SsENMis{} and \SsENQwe{} for Gemma-4, Mistral and Qwen3-VL-32B, respectively), so it cannot exclude a modest harness-specific component. Likewise, F $-$ N directly identifies an effect of rejection-status wording, and interpreting that effect specifically as exclusion remains conditional on the equal-attraction clause of Assumption~\ref{ass:channel}. The neutral replacement shows that deletion does not isolate semantic failure content, so B $-$ A is not a pure attraction contrast. Failure type is nearly confounded with app on this benchmark. The same-harness reconstruction passes the preregistered $\pm0.10$ equivalence margin, but not $\pm0.05$, and the registration states this margin without a separate substantive justification.

\paragraph{Models and serving.} Three open 24--32B models and one 30B instruct/thinking pair, and no proprietary model. Measured rates include serving nondeterminism, and the models differ markedly from one another, so per-model results matter more than pooled ones. Provider routing was not pinned for the main state interventions. Conditions received similar provider mixtures; key earlier contrasts and the post hoc within-provider F $-$ N check have the same direction in their larger groups (App.~\ref{app:protocol}), but smaller groups are unstable and other round-4 contrasts lack provider-specific replication. Only the regime comparison was pinned.

\paragraph{Annotation.} All annotators are LLMs and the codebook was refined on the coded replies. Human annotation of a held-out sample is pending.

\section*{Ethical Considerations}
The study measures optimizer behaviour on synthetic phone screens, and no personal data and no human subjects are involved. Models were accessed through a commercial API within their terms of use. Lower re-proposal makes automated self-improvement cheaper, which inherits the oversight concerns of such loops. An LLM (the authoring assistant) served as one annotator and assisted with code and writing. All numbers are generated from logged data by scripts released with the paper.

\bibliography{refs}

\appendix
\section{Proofs and Estimators}\label{app:proofs}

\paragraph{Identity~\ref{prop:decomp}.} Write $W_{\mathrm{in}}=\sum_{e\in X_s}w_s(e)$ and $W_{\mathrm{out}}=\sum_{e\in M_s\setminus X_s}w_s(e)$. Both sums are positive under Assumption~\ref{ass:choice}, and
\[\beta(s)=\frac{W_{\mathrm{in}} e^{-\lambda_s}}{W_{\mathrm{in}} e^{-\lambda_s}+W_{\mathrm{out}}},\]
so $\operatorname{logit}\beta(s)=\log W_{\mathrm{in}}-\log W_{\mathrm{out}}-\lambda_s$.
The reference has $\operatorname{logit}\beta^{U}(s)=\log|X_s|-\log|M_s\setminus X_s|$. Subtracting,
\[\alpha(s)=\log\frac{W_{\mathrm{in}}}{|X_s|}-\log\frac{W_{\mathrm{out}}}{|M_s\setminus X_s|}-\lambda_s,\]
which is $a(s)-\lambda_s$ because $W_{\mathrm{in}}/|X_s|=\bar w_s(X_s)$ and $W_{\mathrm{out}}/|M_s\setminus X_s|=\bar w_s(M_s\setminus X_s)$. A uniform proposer has $w_s$ constant and $\lambda_s=0$, hence $\alpha=0$. Any $p_s$ with $0<\beta(s)<1$ admits the representation with $\lambda_s=0$ and $w_s=p_s$; more generally, for any chosen $\lambda_s\ge0$, set $w_s(e)=p_s(e)e^{\lambda_s\mathbf{1}[e\in X_s]}$. Thus the split between $a$ and $\lambda$ is not identified from $p_s$ alone, which is why Proposition~\ref{prop:ident} needs Assumption~\ref{ass:channel}.

\paragraph{Identity~\ref{prop:probe}.} By Definition~\ref{def:alpha}, $\operatorname{logit}\beta(s)=\alpha(s)+\operatorname{logit}\beta^{U}(s)$ at every eligible state. Apply Identity~\ref{prop:decomp} at $s$ and $s_0$ and subtract to obtain the enrichment identity; adding $\operatorname{logit}\beta^{U}(s)-\operatorname{logit}\beta^{U}(s_0)$ to both sides gives the raw transport identity. When $\beta^{U}(s)=\beta^{U}(s_0)$ the last bracket vanishes, and since logit is increasing, $\beta(s)\le\beta(s_0)$ holds exactly when $a(s)-a(s_0)\le\lambda_s-\lambda_{s_0}$.

\paragraph{Proposition~\ref{prop:ident}.} An edit $\phi$ leaves $X_s$ and $M_s$ unchanged, so $\beta^{U}(\phi(s))=\beta^{U}(s)$ and the log odds ratio $\operatorname{logit}\beta(\phi(s))-\operatorname{logit}\beta(s)$ equals $\alpha(\phi(s))-\alpha(s)$, which by Identity~\ref{prop:decomp} is $[a(\phi(s))-a(s)]-[\lambda_{\phi(s)}-\lambda_s]$. For an evidence edit the second bracket is zero. For two sentence edits $\phi_1,\phi_2$ listing the same IDs at the same position, both leave $X_s$ and $M_s$ unchanged, so the same argument with $\phi_2(s)$ in place of $s$ gives $[a(\phi_1(s))-a(\phi_2(s))]-[\lambda_{\phi_1(s)}-\lambda_{\phi_2(s)}]$, and the first bracket is zero by Assumption~\ref{ass:channel}.

\paragraph{No-history baseline.} In $s^{\varnothing}$ no edit is displayed as rejected, so under Assumption~\ref{ass:choice} the exclusion term multiplies nothing and $p_{s^{\varnothing}}(e)\propto w_{s^{\varnothing}}(e)$. The menu is still $M_s$, and the source-state set $X_s$ satisfies $0<|X_s|<|M_s|$, so the computation of Identity~\ref{prop:decomp} for the target set $X_s$, with no exclusion factor, gives $\alpha_{X_s}(s^{\varnothing})=a_{X_s}(s^{\varnothing})$. For the original state, $\alpha_{X_s}(s)=a_{X_s}(s)-\lambda_s$. Hence $\alpha_{X_s}(s)-\alpha_{X_s}(s^{\varnothing})=[a_{X_s}(s)-a_{X_s}(s^{\varnothing})]-\lambda_s$: the net effect of adding the ordinary history on the mass assigned to $X_s$. The contrast identifies $\lambda_s$ only if the history leaves $a_{X_s}$ unchanged; we do not assume that, and report the net effect.

\paragraph{Estimators.} \emph{Rejection-enrichment log-odds.} For steps $t$ with outcomes $B_t$ and offsets $o_t=\operatorname{logit}\beta^{U}(s_t)$, $\hat\alpha$ maximises $\sum_t[B_t(\alpha+o_t)-\log(1+e^{\alpha+o_t})]$, a concave function; we use damped Newton steps. It is the maximum-likelihood estimator if $\alpha$ is constant over the steps and the replies are independent given their states. Steps of one run are not independent, so we treat it as a working likelihood, solving $\sum_t[B_t-\sigma(\hat\alpha+o_t)]=0$, and take CIs from the cluster bootstrap. When $\alpha$ varies across steps, $\hat\alpha$ is a weighted summary. \emph{Probe bound.} With no re-proposal in cells with offsets $o_i$, the bound is the largest $\alpha$ with $\prod_i(1-\sigma(\alpha+o_i))\ge0.05$, exact if the cells are independent with a common $\alpha$; with $k>0$ re-proposals, $P(\sum_iB_i\le k)$ under the Poisson-binomial distribution replaces the product. \emph{State edits.} With $a_s,b_s$ the re-proposals and non-re-proposals of state $s$ under the edit and $c_s,d_s$ under the comparison, the Mantel--Haenszel estimate is $\log\bigl(\sum_s a_sd_s/n_s\bigr)-\log\bigl(\sum_s b_sc_s/n_s\bigr)$, consistent for a common odds ratio as the number of states grows with few calls per state, assuming independent calls within a state and condition; when the state-level log odds ratios differ, it estimates a weighted average of them.

\section{Registration, Claims and All Registered Tests}\label{app:prereg}

Each confirmatory experiment has a JSON registration file (\texttt{exp/prereg/}). The anonymous Data supplement contains the registration files, per-call ledger and raw replies; the Software supplement contains the analysis scripts. The files carry their own UTC timestamps; they were not deposited with an external registry, and the order of registration and data collection can be checked against the per-call ledger (Table~\ref{tab:prereg}). Each file states the question, hypothesis, falsifier, primary endpoint, models, seeds, sampling, unit of analysis, exclusions, test, CI, multiple-testing family, stopping rule and budget. Ten pilot calls preceded registration; they were used only to check the parser and are excluded from every analysis. The registered analyses are differences in rates; the log-odds analyses of \S\ref{sec:model} were added afterwards, at no cost, from the same logged data (\texttt{analysis/model.py}).

\begin{table}[h]
\centering\scriptsize\setlength{\tabcolsep}{3pt}
\resizebox{\columnwidth}{!}{\begin{tabular}{@{}lllr@{}}
\toprule
Registration file & Registered (UTC) & First paid call (UTC) & USD \\
\midrule
\texttt{VRR\_CONFIRM.json} & 2026-10-04 06:49:05 & 2026-10-04 06:49:43 & 0.393 \\
\texttt{VRR\_INTERVENE.json} & 2026-10-04 06:49:05 & 2026-10-04 07:15:52 & 0.375 \\
\texttt{VRR\_REASONING.json} & 2026-10-04 06:49:05 & 2026-10-04 07:17:11 & 1.208 \\
\texttt{VRR\_REASONING\_AMEND1.json} & 2026-10-04 08:34:01 & 2026-10-04 08:34:25 & 0.144 \\
\texttt{VRR\_QUAL.json} & 2026-10-04 06:49:05 & 2026-10-04 07:20:41 & 0.014 \\
\texttt{VRR\_R2\_REASONING.json} & 2026-10-04 10:10:24 & 2026-10-04 10:12:04 & 0.282 \\
\texttt{VRR\_R2\_STATEPROBE.json} & 2026-10-04 10:10:24 & 2026-10-04 10:12:05 & 0.131 \\
\texttt{VRR\_R2\_SEEDS.json} & 2026-10-04 10:10:24 & 2026-10-04 10:12:04 & 0.087 \\
\texttt{VRR\_R2\_QUAL.json} & 2026-10-04 10:10:24 & 2026-10-04 10:12:04 & 0.029 \\
\texttt{VRR\_R3.json} & 2026-10-04 14:06:25 & 2026-10-04 14:07:34 & 0.029 \\
\texttt{VRR\_R4.json} (VRR-R4) & 2026-10-07 15:01:24 & 2026-10-07 15:01:38 & 0.973 \\
\texttt{VRR\_R4.json} (VRR-R4-REGIME) & 2026-10-07 15:01:24 & 2026-10-07 15:46:36 & 1.215 \\
\bottomrule
\end{tabular}
}
\caption{Registration vs.\ first paid call (UTC) and spend per experiment.}
\label{tab:prereg}
\end{table}

\begin{table*}[t]
\centering\scriptsize\setlength{\tabcolsep}{3pt}
\begin{tabularx}{\textwidth}{@{}>{\raggedright\arraybackslash}p{4.6cm}>{\raggedright\arraybackslash}p{1.1cm}>{\raggedright\arraybackslash}X>{\raggedright\arraybackslash}p{3.3cm}>{\raggedright\arraybackslash}p{2.2cm}@{}}
\toprule
Claim & Where & Evidence & Falsifier & Status \\
\midrule
Loop re-proposal exceeds clean-probe re-proposal & \S\ref{sec:res-transport} & H1: $\bar\Delta=$\DeltaMacro{} \DeltaMacroCI; H1b: \HoneB/6 cells & CI includes 0; fewer than 4 cells & registered, supported \\
M and A satisfy the preregistered $\pm0.10$ equivalence criterion & \S\ref{sec:res-transport} & H15: M $-$ A \RfEq{} \RfEqCI{} (90\%), 360 fresh states & CI outside $\pm0.10$ & registered, passes $\pm0.10$, not $\pm0.05$ \\
Ordinary history changes re-proposal relative to matched no-history & \S\ref{sec:res-transport} & H14a: A $-$ H \RfAH{} \RfAHCI & CI includes 0 & registered, supported pooled (decrease); reversed for Mistral \\
Failure deletion lowers re-proposal & \S\ref{sec:res-failure} & H2a \Htwoa; H13a \RfBa{} \RfBaCI & CI includes 0 & registered, supported (both rounds) \\
Failure content beyond approximate text-bulk control & \S\ref{sec:res-failure} & H13b: R $-$ A \RfRa{} \RfRaCI & Holm $p\ge0.05$ & registered, not supported (Holm $p$ \RfRaP) \\
Choice follows shuffled failure attachment & \S\ref{sec:res-failure} & H13c: J $-$ K \RfJK; H13d: follow \RfFolA{}$\to$\RfFolS & no difference & registered, not supported \\
Status wording beyond neutral same-ID list & \S\ref{sec:res-signal} & H14b: F $-$ N \RfFN{} \RfFNCI & CI includes 0 & registered, supported in every model \\
Prohibition beyond status wording & \S\ref{sec:res-signal} & H13e: P $-$ F \RfPF{} \RfPFCI; log OR \RfPFLo{} \RfPFLoCI & Holm $p\ge0.05$ & registered rate test not supported (Holm $p$ \RfPFP) \\
Instruct vs.\ thinking, same token cap & \S\ref{sec:res-boundary} & H4a \Hfoura; pinned H16 \Rg{} \RgCI & CI includes 0 & registered, supported; endpoint difference \\
\bottomrule
\end{tabularx}
\caption{\textbf{Claims, evidence and falsifiers.} Registered tests are differences in rates (sign-flip tests, Holm within their family). H13--H16 were registered after the framework of \S\ref{sec:model}.}
\label{tab:claims}
\end{table*}

\paragraph{Amendment to VRR-P1.} The reasoning experiment's registered stop rule (\$1.00) was checked before each 30-iteration run, not before each call. Thinking calls cost about \$0.013 each, roughly ten times the estimate, so the third thinking loop ended at \$1.20. At that point the loop-level counts had been printed, and no cross-evaluation call had been made. The amendment (\texttt{VRR\_REASONING\_AMEND1.json}) reduced the matched-state cross-evaluation from 30+30 to 15+15 states, dropped the thinking-model probe, and moved the stop rule (\$1.75) to every call. This test (H3: \CrossDiff{} \CrossCI{} on \CrossPaired{} paired states, unequal budgets) is superseded by the fresh test with one token cap in \S\ref{sec:res-boundary}.

\paragraph{Round 2.} A second set of registrations (\texttt{VRR\_R2\_*.json}) was written in response to a review and before any of its calls: the fresh three-arm regime test (H4a, H4b), the same-harness probe and the neutral-list control (H5, H6a, H6b), five more seeds per condition on 30 states, and the blinded re-annotation. H5 and H6a/b use the 120 sampled states. They were registered after the results of conditions A--E on those states had been seen, and only conditions M and N were new; we therefore call them preregistered follow-up tests, not independent confirmations. In the blinded re-annotation, gpt-oss-120b's first reply hit its 600-token limit in 11 of 60 cases and was re-queried with 2000 tokens.

\paragraph{Round 3 (stopped).} \texttt{VRR\_R3.json} registered a re-test of H2a, H2b and H6a on 120 fresh states, an equivalence test of the same-harness probe (TOST, margin $\pm0.10$), an approximately length-matched neutral replacement of the failures and a shuffle of failures across entries. The run stopped after \RthreeCalls{} of about \RthreePlanned{} planned calls (\RthreeStates{} \RthreeModels{} states; \$\SpentRthree), when the API key, which is shared with other projects, reached its spending limit. These calls cover too few states for any registered test, so we do not analyse them; the raw replies are released and the spend is included in all totals. Its design was carried into round 4.

\paragraph{Round 4 (fresh states).} \texttt{VRR\_R4.json} was registered after the framework of \S\ref{sec:model} and before any call of the round. It sampled 360 fresh eligible states (60 per model and catalog, none used before), ran conditions A, H, B, R, S, J, K, T, N, F, P, E and M on 120 of them and A and M on the other 240, and pinned the regime comparison to one provider (VRR-R4-REGIME). The first two states were a pilot, read reply by reply; no rendering fault was found, so they are part of the data. Table~\ref{tab:r4} lists every registered test. Of \RfCalls{} calls, a share were served from the cache because the prompt was identical to another condition's (for example K equals A in states whose shown entries are all rejected).

\begin{table*}[t]
\centering\scriptsize\setlength{\tabcolsep}{3pt}
\begin{tabular}{@{}llrllll@{}}
\toprule
Test & Contrast & $n$ & Rate difference [95\% CI] & Hierarchical CI & log OR [95\% CI] & $p_\mathrm{Holm}$ \\
\midrule
H13a & B $-$ A $<0$ & 120 & $-$0.17 [$-$0.23, $-$0.10] & [$-$0.25, $-$0.08] & $-$1.49 [$-$2.21, $-$0.94] & $<$0.001 \\
H13b & R $-$ A $<0$ & 120 & $-$0.07 [$-$0.13, $-$0.01] & [$-$0.15, 0.02] & $-$0.56 [$-$1.16, $-$0.04] & 0.071 \\
H13c & J $-$ K $<0$ & 65 & $+$0.07 [$-$0.02, 0.16] & [$-$0.03, 0.18] & $+$0.53 [$-$0.10, 1.24] & 0.928 \\
H13d & follow S $>$ A & 102 & $+$0.02 [$-$0.01, 0.05] & -- & -- & 0.425 \\
H13e & P $-$ F $<0$ & 120 & $-$0.05 [$-$0.10, $-$0.01] & [$-$0.11, 0.00] & $-$1.15 [$-$2.40, $-$0.31] & 0.054 \\
H13f & E $-$ A $<0$ & 120 & $-$0.27 [$-$0.32, $-$0.22] & [$-$0.34, $-$0.20] & $-$3.61 [$-$5.05, $-$2.89] & $<$0.001 \\
\midrule
H14a & A $-$ H & 120 & $-$0.21 [$-$0.30, $-$0.11] & [$-$0.34, $-$0.07] & $-$0.68 [$-$1.06, $-$0.34] & 0.002 \\
H14b & F $-$ N & 120 & $-$0.16 [$-$0.22, $-$0.09] & [$-$0.25, $-$0.06] & $-$1.44 [$-$2.30, $-$0.81] & $<$0.001 \\
H14c & R $-$ B & 120 & $+$0.10 [0.05, 0.14] & [0.04, 0.16] & $+$1.25 [0.64, 2.01] & 0.002 \\
\midrule
H15 & M $-$ A, equivalence $\pm0.10$ & 360 & $-$0.04 [$-$0.07, $-$0.01] (90\%) & [$-$0.08, 0.00] & $-$0.36 & equivalent \\
H16 & thinking $-$ instruct, pinned & 40 & $-$0.33 [$-$0.43, $-$0.22] & -- & -- & $<$0.001 \\
\bottomrule
\end{tabular}

\caption{\textbf{Round-4 registered tests on fresh states.} One-sided tests H13a--f (Holm over six), two-sided H14a--c (Holm over three); stratified and hierarchical (run, then state) bootstrap CIs; Mantel--Haenszel log odds ratios. H15: equivalence of M $-$ A within $\pm0.10$ (90\% CI over 360 states). H16: provider-pinned regime comparison.}
\label{tab:r4}
\end{table*}

\begin{table}[h]
\centering\small\setlength{\tabcolsep}{2.4pt}
\resizebox{\columnwidth}{!}{\begin{tabular}{@{}llrlr@{}}
\toprule
Test & Contrast & $n$ & Effect [95\% CI] & $p_\mathrm{Holm}$ \\
\midrule
H4a & thinking $-$ instruct (both 4000 tok.) & 40 states & $-$0.40 [$-$0.55, $-$0.25] & $<$0.001 \\
H4b & instruct 4000 $-$ 200 tok. & 36 states & $+$0.06 [$-$0.06, 0.17] & 0.689 \\
H5 & same-harness probe $-$ loop state & 117 states & $-$0.05 [$-$0.11, 0.01] & 0.177 \\
H6a & DO-NOT $-$ neutral list & 120 states & $-$0.23 [$-$0.29, $-$0.18] & $<$0.001 \\
H6b & neutral list $-$ unchanged & 117 states & $-$0.06 [$-$0.13, 0.01] & 0.177 \\
\bottomrule
\end{tabular}
}
\caption{\textbf{Round-2 registered tests.} Paired differences in re-proposal rate over states; sign-flip tests, Holm within each registration (H4a/b; H5, H6a/b).}
\label{tab:r2}
\end{table}

\paragraph{Outcomes of all registered hypotheses.} H1 (macro gap $>0$): supported. H1b (positive in $\ge$4 of 6 cells): supported, \HoneB/6. H2a ($B-A<0$): supported. H2b ($E-A<0$): supported, \Htwob{} \HtwobCI. H2c ($C-A$, two-sided): CI includes 0. H2d ($D-A$, two-sided): negative, $p_\mathrm{Holm}$ \HtwodP. H3: supported under the amendment; superseded by H4a. H4a: supported. H4b: CI includes 0. H5: CI includes 0. H6a: supported. H6b: CI includes 0. QUAL: $\kappa$ reported as registered (first round $\kappa=$\Kappa; blinded $\kappa=$\KappaBlind). VRR-R3: not tested (run stopped). Round 4 (Table~\ref{tab:r4}): H13a and H13f supported; H13b, H13c, H13d and H13e not supported after Holm; H14a, H14b and H14c supported; H15 passes the $\pm0.10$ margin but not $\pm0.05$; H16 supported.

\paragraph{Heterogeneity and sensitivity of the registered endpoint.} The six cells share three models, one candidate set and one task, so we do not summarise them as a random sample of cells. Per model, averaged over catalogs, $\Delta$ is \DeltaGem{} \DeltaGemCI{} (Gemma-4), \DeltaMis{} \DeltaMisCI{} (Mistral) and \DeltaQwe{} \DeltaQweCI{} (Qwen3-VL-32B). The difference between catalogs (catalog~B minus catalog~A) includes 0 for Gemma-4 and Mistral and is positive for Qwen3-VL-32B (\CatDiffQwe{} \CatDiffQweCI; \CatStratQweCI{} with the run bootstrap stratified by view; \QweCatAdj{} \QweCatAdjCI{} with view and $|X_t|$ fixed). Subtracting the exact 95\% upper bound of each probe rate from the lower bound of the loop CI leaves the gap positive in all cells except Gemma-4 and Qwen3-VL-32B on catalog~A (\ConsNeg{} of 6), where loop rates are lowest.

\section{Protocol, Models and Costs}\label{app:protocol}

\paragraph{Models.} {\sloppy Proposers via OpenRouter: \texttt{google/}\allowbreak\texttt{gemma-4-26b-a4b-it}, \texttt{mistralai/}\allowbreak\texttt{mistral-small-3.2-24b-instruct}, \texttt{qwen/}\allowbreak\texttt{qwen3-vl-32b-instruct}, \texttt{qwen/}\allowbreak\texttt{qwen3-30b-a3b-instruct-2507}, \texttt{qwen/}\allowbreak\texttt{qwen3-30b-a3b-thinking-2507}. Annotators: \texttt{deepseek/}\allowbreak\texttt{deepseek-chat-v3.1}, \texttt{openai/}\allowbreak\texttt{gpt-oss-120b}. Agent: \texttt{qwen/}\allowbreak\texttt{qwen3-vl-8b-instruct}. Prices were fetched from the provider API before the first call (\texttt{exp/}\allowbreak\texttt{configs/pricing.json}). Temperature is 0 throughout; the state edits also set provider seeds. The loop prompt asks for at most two sentences of reasoning and a final line \texttt{CHOICE: <id>}.\par}

\paragraph{Serving providers.} OpenRouter routed every call to one of the providers that serve the model; we did not pin providers. The ledger records the serving provider of each call (Table~\ref{tab:providers}). Within a model, re-proposal rates differ between providers; in condition A, for example, Mistral re-proposed in \MisDIAB{} of \MisDIAN{} valid DeepInfra replies and \MisVenAB{} of \MisVenAN{} valid Venice replies. The provider mix was similar across conditions (Mistral: Venice served \MisVeniceMin--\MisVeniceMax{} of 120 calls per condition), and B $<$ A and E $<$ A hold within each of the \ProvMainN{} model--provider combinations with at least 25 valid replies per condition. These checks reduce concern about routing but cannot establish its absence as a confounder. Round 4 also routed freely, except for the regime rerun (VRR-R4-REGIME), whose 240 calls were all served by the pinned provider.

\begin{table}[h]
\centering\scriptsize\setlength{\tabcolsep}{2.5pt}
\resizebox{\columnwidth}{!}{\begin{tabular}{@{}lrrl@{}}
\toprule
Model & Calls & Providers & Largest shares \\
\midrule
Gemma-4-26B-A4B & 3295 & 13 & NextBit 34\%, Cloudflare 22\%, DekaLLM 11\%, \ldots \\
Mistral-Small-3.2-24B & 2962 & 3 & Venice 57\%, DeepInfra 34\%, Parasail 9\% \\
Qwen3-VL-32B & 3021 & 1 & Alibaba 100\% \\
Qwen3-30B-A3B-Instruct & 347 & 6 & Alibaba 39\%, StreamLake 32\%, SiliconFlow 10\%, \ldots \\
Qwen3-30B-A3B-Thinking & 256 & 1 & Alibaba 100\% \\
Qwen3-VL-8B (agent) & 2640 & 3 & PrimeIntellect 44\%, Alibaba 43\%, Parasail 13\% \\
DeepSeek-V3.1 (annotator) & 120 & 4 & DeepInfra 52\%, AtlasCloud 43\%, SiliconFlow 4\%, \ldots \\
gpt-oss-120b (annotator) & 71 & 6 & DigitalOcean 92\%, CoreWeave 3\%, Amazon Bedrock 1\%, \ldots \\
\bottomrule
\end{tabular}
}
\caption{Serving providers per model over all paid calls except the pilot (from the per-call ledger).}
\label{tab:providers}
\end{table}

\paragraph{Post hoc provider check for F $-$ N.} Table~\ref{tab:provider-fn} compares valid F and N replies only where both conditions reached the same state through the same provider; multiple calls within a state--condition--provider are averaged before the state contrast. All four model--provider groups with at least 25 valid calls under each condition have negative estimates. The Gemma-4/Io Net group has a positive estimate from five paired states; several other groups have wide intervals. This check is descriptive because provider assignment was not controlled, and one additional group with fewer than five paired states is omitted from the table.

\begin{table*}[t]
\centering\small
\resizebox{\textwidth}{!}{\begin{tabular}{@{}llrrl@{}}
\toprule
Model & Provider & F/N calls & Paired states & F $-$ N [95\% CI] \\
\midrule
Gemma-4 & Cloudflare & 26/26 & 9 & $-$0.15 [$-$0.33, 0.00] \\
Gemma-4 & Io Net & 14/13 & 5 & $+$0.07 [0.00, 0.20] \\
Gemma-4 & Makora & 18/18 & 7 & $-$0.33 [$-$0.67, $-$0.05] \\
Gemma-4 & NextBit & 47/45 & 16 & $-$0.11 [$-$0.28, 0.01] \\
Mistral & DeepInfra & 18/11 & 9 & $-$0.11 [$-$0.39, 0.11] \\
Mistral & Parasail & 15/14 & 7 & $-$0.21 [$-$0.79, 0.43] \\
Mistral & Venice & 78/81 & 33 & $-$0.26 [$-$0.43, $-$0.09] \\
Qwen3-VL & Alibaba & 117/115 & 40 & $-$0.11 [$-$0.18, $-$0.05] \\
\bottomrule
\end{tabular}
}
\caption{\textbf{Provider sensitivity of F $-$ N (post hoc).} Valid F/N calls are counted within paired states, which have at least one valid call under both conditions from the same provider. Differences average state-level rates within each provider, with catalog-stratified 95\% bootstrap CIs over paired states. Only groups with at least five paired states are shown.}
\label{tab:provider-fn}
\end{table*}

\paragraph{Harness edits.} A reply naming an edit already in $h_s$ is invalid in the loops and counted as $B=0$ in the state edits, where it makes up \HarnRepPtwo\% and \HarnRepRfour\% of valid replies in the two rounds.

\paragraph{Caching.} Every call is stored under the hash of (model, messages, max tokens, temperature, provider, seed), and a repeated identical request returns the stored reply. Agent evaluations are cached per harness, so a harness has one development accuracy. The response cache of an earlier exploratory study on the same benchmark was read (never written) as a fallback; it supplied cached agent evaluations and the base-harness rejections used in the probe.

\paragraph{Budget.} A hard project cap of \$10 applied; the client refuses any call whose estimated cost would exceed the project budget, which was \$3.5 until round 4 and was then raised to the cap with the authors' approval. Each registration also set its own stop rule. Table~\ref{tab:budget} gives the spend per experiment.

\begin{table}[h]
\centering\scriptsize
\begin{tabular}{@{}lr@{}}
\toprule
Experiment & USD \\
\midrule
VRR-P0-1 & 0.3381 \\
VRR-P0-1-probe & 0.0545 \\
VRR-P0-2 & 0.3748 \\
VRR-P1 & 1.1997 \\
VRR-P1-cross & 0.1440 \\
VRR-P1-probe & 0.0085 \\
VRR-QUAL & 0.0141 \\
VRR-R2-QUAL & 0.0290 \\
VRR-R2-REASON & 0.2816 \\
VRR-R2-SEEDS & 0.0875 \\
VRR-R2-SPROBE & 0.1310 \\
VRR-R3 & 0.0288 \\
VRR-R4 & 0.9731 \\
VRR-R4-REGIME & 1.2154 \\
VRR-pilot & 0.0013 \\
\midrule
Total & 4.8815 \\
\bottomrule
\end{tabular}

\caption{Spend per experiment (USD, from the per-call ledger).}
\label{tab:budget}
\end{table}

\section{Verification Log}\label{app:verify}

Every boundary case and unexpected result was traced through the code path and the logged data before it was interpreted. Each entry gives the observation, the check and the outcome.

\begin{itemize}[leftmargin=*,itemsep=2pt,topsep=1pt]
\item \emph{Probe rates of exactly zero.} Checked that the probe's window edits are rejected base evaluations, that the parser accepts both catalogs' IDs, and that the favourite is in the window. Outcome: genuine zeros; reported as exact upper bounds, not as $-\infty$.
\item \emph{Raw rates mislead when $|X_t|$ varies.} Full views show more rejections (mean $|X_t|$ \NxF{} vs.\ \NxW). Outcome: the uniform-menu reference (Definition~\ref{def:alpha}); the FULL $-$ W8 loop-rate difference shrinks from \ViewRaw{} to \ViewAdj{} \ViewAdjCI{} with cell and $|X_t|$ fixed in a linear probability model. Qwen3-VL-32B's catalog effect survives the same adjustment (\QweCatAdj{} \QweCatAdjCI).
\item \emph{Harness or rejected-set size?} Base-harness loop steps have $|X_t|\le3$ for Gemma-4 and Qwen3-VL-32B. Outcome: the harness contrast holds at matched $|X_t|\le3$ (\S\ref{sec:res-transport}).
\item \emph{Depletion of the menu.} Hypothesis: avoidance falls as edits are kept. Test: slope of $\alpha$ in $|h_t|$ with cell intercepts. Outcome: not consistent across models (Gemma-4 \GamGem{} \GamGemCI, Mistral \GamMis{} \GamMisCI, Qwen3-VL-32B \GamQwe{} \GamQweCI); not used as an explanation.
\item \emph{Self-selection of the rejected set.} Every edit of $X_s$ was chosen by the proposer at $h_s$, so H's enrichment of $X_s$ is expected under a stable preference. Null set $O_s$: edits the proposer chose earlier in the run at other harnesses, minus $X_s$ and $h_s$, on the states where $0<|O_s|<|M_s|$ (\SsENGem, \SsENMis{} and \SsENQwe{} states). Under H, $\hat\alpha$ for $X_s$ and $O_s$ is \SsEXGem{} and \SsEOGem{} (Gemma-4), \SsEXMis{} and \SsEOMis{} (Mistral), \SsEXQwe{} and \SsEOQwe{} (Qwen3-VL-32B); the excess of $X_s$ is \SsEExGem{} \SsEExGemCI, \SsEExMis{} \SsEExMisCI{} and \SsEExQwe{} \SsEExQweCI. With $O_s$ taken from the whole run (\SsRNGem, \SsRNMis{} and \SsRNQwe{} states) the excess is \SsRExGem{} \SsRExGemCI, \SsRExMis{} \SsRExMisCI{} and \SsRExQwe{} \SsRExQweCI. Outcome: no detectable harness-specific excess; \S\ref{sec:res-transport} describes the enrichment as consistent with self-selection (post hoc, paired bootstrap over states, \texttt{analysis/r4.py}).
\item \emph{Base-harness zero-history favourites.} Hypothesis: loops fill $X_t$ with the proposer's zero-history favourites at the base harness. Outcome: rejected; the zero-history favourites mostly remain available outside $X_t\cup h_t$, and the edit Mistral re-proposes most often (E33) is not among them.
\item \emph{Failure-target availability (exploratory; used in \S\ref{sec:res-transport}).} Hypothesis: avoidance fails when the edits a model chooses in its probe (its two most frequent probe choices per catalog) are in $X_t$. Outcome: for Gemma-4, $\hat\alpha=$\TOutGem{} \TOutGemCI{} without and \TInGem{} \TInGemCI{} with a target in $X_t$; for Qwen3-VL-32B, \TOutQwe{} \TOutQweCI{} and \TInQwe{} \TInQweCI; for Mistral, \TOutMis{} \TOutMisCI{} and \TInMis{} \TInMisCI. Supportive for two models, not for Mistral, which is attracted either way; the targets were chosen after seeing the probe data.
\item \emph{Low agreement among seeded temperature-0 calls} (\SeedAgreeMin--\SeedAgreeMax{} of state--condition triples agree). Check: provider routing. Outcome: pairs served by the same provider agree no more often than pairs served by different providers (\SeedSame{} vs.\ \SeedDiff); the state rate is treated as a Monte Carlo estimate. With eight calls on 30 states, B $-$ A is \SBeight{} \SBeightCI{} (three calls: \SBthree) and E $-$ A \SEeight{} \SEeightCI{} (\SEthree).
\item \emph{Dependence among states from one run.} The 120 states come from \HBruns{} source runs. Outcome: a hierarchical bootstrap that resamples runs within model and catalog and then states within runs gives log odds ratios with CIs \HBoBCI{} (B $-$ A), \HBoCCI{} (C $-$ A), \HBoDCI{} (D $-$ A), \HBoECI{} (E $-$ A), \HBoNCI{} (N $-$ A), \HBoMCI{} (M $-$ A) and \HBoENCI{} (E $-$ N), and rate differences \HBdBCI, \HBdCCI, \HBdDCI, \HBdECI, \HBdNCI, \HBdMCI{} and \HBdENCI. Every conclusion drawn from the state-level CIs holds.
\item \emph{The neutral replacement (R) restores most of the rate that deletion (B) removes.} This was the most consequential unexpected result. We rebuilt R prompts and checked that each failure segment was replaced in place, with the same entry layout and a length within 13 characters, and that no failure text survived. Outcome: R is constructed as registered. The neutral replacement restores most of the deletion effect for Mistral; Gemma-4 shows little effect from either deletion or neutral replacement; Qwen3-VL-32B retains a substantial effect under neutral replacement, consistent with a content-sensitive component.
\item \emph{A strict type swap is rarely possible.} Only 73 of 2{,}294 shown ordinal failures have an unshown non-ordinal failure from the same evaluation and app within 10 characters, because failure type and app nearly coincide on this benchmark (mail-list and file-grid failures are ordinal). Outcome: T is descriptive (\RfWithT{} states).
\item \emph{The history raises re-proposal for Mistral.} A $-$ H is negative for Gemma-4 and Qwen3-VL-32B but positive for Mistral (\RpAHMis{} \RpAHMisCI). Checked that H prompts keep the harness line, catalog and question and differ from A only in the history block. Outcome: reported as a model difference, not pooled away.
\item \emph{Cached replies in round 4.} 900 of the round's calls were served from the cache because their prompt equalled another condition's prompt for the same state and seed (for example K equals A when every shown entry is rejected, and M equals A when the history holds only rejected entries). These are genuine replies to the identical request and are counted once per condition.
\item \emph{Iteration cost versus visible re-proposal.} The \CostPctMin--\CostPctMax\% figure counts duplicate pairs ($D_t$), which Table~\ref{tab:p01} does not show. Recomputed from the loop logs: every step with $B_t=1$ has $D_t=1$, and steps with $D_t=1$, $B_t=0$ (\CostHidGem{} Gemma-4, \CostHidMis{} Mistral, \CostHidQwe{} Qwen3-VL-32B) occur only under the W8 view, where the earlier attempt has left the window. Denominators are all valid loop steps, not only eligible ones (\texttt{analysis/model.py}, COST).
\item \emph{Condition-D fallback.} Recomputed from the stored states: \DfbShare\% of replaced entries used base-harness failures, in \DfbStates{} of 120 states; on the \DownN{} states without fallback, D $-$ A is \Down{} \DownCI. Only \DstrictN{} state has replacements that come from the same evaluation and app and match the original failures in length within 10 characters, so a strict type swap needs newly constructed prompts.
\item \emph{Cached replies in the regime test.} Matching replies against the ledger: \CachedN{} of \CachedTot{} replies were served from the cache because the request equalled an earlier one; \CachedOwn{} are the original loop replies (I-200 on instruct-loop states, T-4000 on thinking-loop states). States were sampled without regard to these replies.
\item \emph{Invalid replies.} \InvTotal{} of 1080 loop steps were invalid: \InvTruncGem{} Gemma-4 and \InvTruncQwe{} Qwen3-VL-32B replies hit the 200-token limit before \texttt{CHOICE:}, Mistral named an edit already in the harness \InvHarnMis{} times, and one catalog-B reply named a code that does not exist. Counting them as compliant or as re-proposals gives $\bar\Delta=$\DeltaInvZero{} \DeltaInvZeroCI{} and \DeltaInvOne{} \DeltaInvOneCI.
\item \emph{Formulas and estimators.} Identity~\ref{prop:decomp} was checked numerically on 10{,}000 random weightings (agreement to numerical precision). Every reported $\hat\alpha$ satisfies its score equation, and the Mantel--Haenszel estimator recovered a known common log odds ratio in simulated sparse strata of three calls. No eligible state has $X_s=M_s$, so $\beta^{U}\in(0,1)$ throughout. Replies naming an edit already in the harness are invalid in the loops and counted as $B=0$ in the state edits (\HarnRepPtwo\% and \HarnRepRfour\% of valid replies).
\item \emph{Divergent log-odds fits.} An undamped Newton iteration diverged for Mistral, whose rate is far above the reference. Outcome: damped steps with a log-likelihood check; all fits converge and the intercept-only estimates match a grid search of the likelihood.
\item \emph{Earlier replay disagreed for Mistral.} On violation-selected states, removing failures did not lower Mistral's rate; on random states it does (\LoBMis{} \LoBMisCI). Outcome: attributed to selection on a prior violation (App.~\ref{app:earlier}).
\item \emph{Pipeline faults.} A non-reentrant lock in the budget check hung the first pilot (fixed before registration; no data affected). The first-round second annotator used the state summary against the codebook; we report that round rather than re-prompting, and blinded the second round.
\end{itemize}

\section{Supplementary Results}\label{app:supp}

\begin{figure}[h]
\centering
\includegraphics[width=\columnwidth]{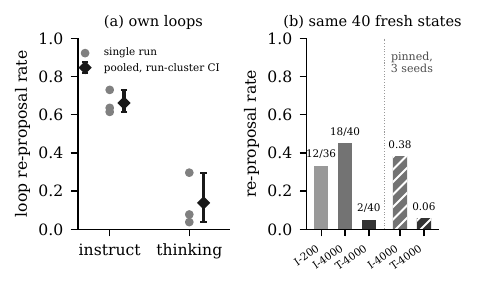}
\caption{\textbf{Instruct vs.\ thinking (Qwen3-30B-A3B-2507).} (a) Re-proposal rate in each model's own loops. (b) The same \ReasonN{} fresh states answered by the instruct model with a 200- or 4000-token cap and by the thinking model with a 4000-token cap: unpinned routing, one call per state (left), and both endpoints pinned to one provider with three seeds (hatched).}
\label{fig:reasoning}
\end{figure}

\begin{figure}[h]
\centering
\includegraphics[width=\columnwidth]{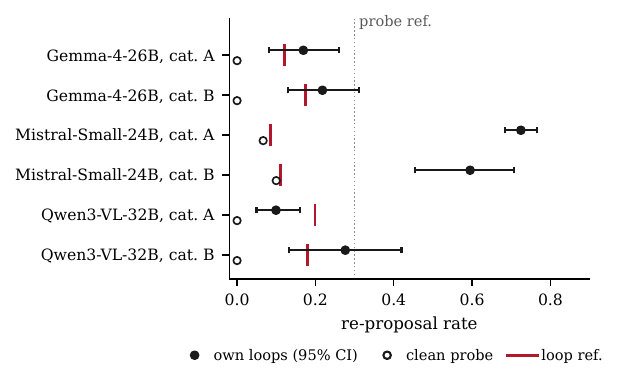}
\caption{\textbf{Raw re-proposal rates against the uniform-menu reference.} Filled: own loops, run-cluster 95\% CIs; open: clean probe. Red ticks: each loop's reference rate $|X_t|/(40-|h_t|)$; dotted line: the probe's reference, $L/40$.}
\label{fig:rates}
\end{figure}

\begin{figure}[h]
\centering
\includegraphics[width=\columnwidth]{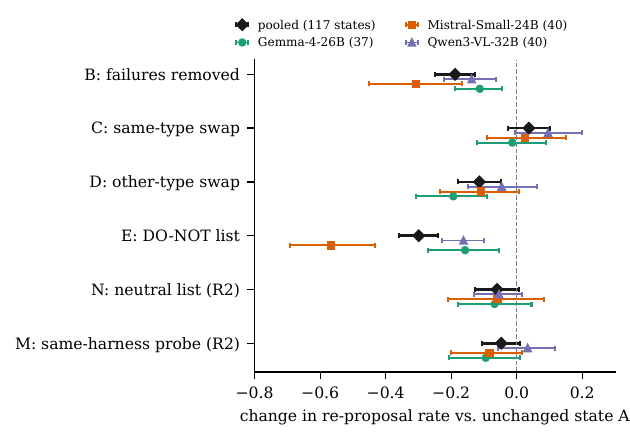}
\caption{\textbf{State edits as differences in rate}, pooled and per model (state counts in the legend), 95\% bootstrap CIs over states.}
\label{fig:interventions}
\end{figure}

\begin{figure}[h]
\centering
\includegraphics[width=\columnwidth]{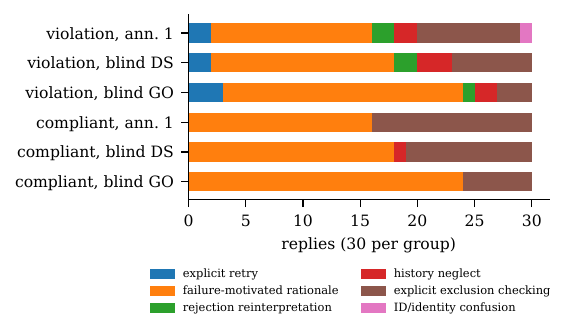}
\caption{\textbf{Stated reasons} for re-proposals and avoiding choices: annotator~1 and the two blinded LLM annotators (30 replies per group; blinded pair $\kappa=$\KappaBlind). ``Failure-motivated rationale'' abbreviates the preregistered code \emph{failure-target attraction} (App.~\ref{app:codebook}), which describes the stated rationale, not a causal mechanism.}
\label{fig:taxonomy}
\end{figure}

Table~\ref{tab:r4models} gives the round-4 contrasts per model, Fig.~\ref{fig:reasoning} the regime comparison and Fig.~\ref{fig:taxonomy} the code frequencies of \S\ref{sec:res-boundary}. Fig.~\ref{fig:rates} shows the raw rates behind Table~\ref{tab:p01}, and Fig.~\ref{fig:interventions} the registered rate differences behind Table~\ref{tab:p02}. Table~\ref{tab:permodel} gives the per-model registered contrasts and Table~\ref{tab:runs} every loop. In the regime study, on states sampled from the instruct model's loops the thinking model's rate is lower by \CrossFromIns, and on states from its own loops by \CrossFromThk; the instruct model gave \InvIns{} invalid replies on eligible loop steps and the thinking model \InvThk.

\paragraph{Reconstruction sensitivity.} Table~\ref{tab:history-sensitivity} separates states whose A and M prompts are identical from those whose prompts change. The first \RfSameN{} states contribute zero by construction to M $-$ A. On the remaining \RfChangedN{} states, M $-$ A is \RfChanged{} (90\% CI \RfChangedCI; run-hierarchical \RfChangedCIh). This split and its per-model estimates are post hoc; the preregistered H15 result uses all 360 states.

\begin{table*}[t]
\centering\small
\begin{tabular}{@{}lrll@{}}
\toprule
State subset & $n$ & M $-$ A [90\% CI] & Run-hierarchical [90\% CI] \\
\midrule
A and M identical & 117 & $+$0.000 [0.000, 0.000] & -- \\
A and M differ & 243 & $-$0.057 [$-$0.097, $-$0.016] & [$-$0.113, 0.003] \\
\quad Gemma-4-26B-A4B & 98 & $+$0.005 [$-$0.060, 0.068] & -- \\
\quad Mistral-Small-3.2-24B & 71 & $-$0.099 [$-$0.178, $-$0.019] & -- \\
\quad Qwen3-VL-32B & 74 & $-$0.099 [$-$0.164, $-$0.036] & -- \\
\bottomrule
\end{tabular}

\caption{\textbf{Same-harness reconstruction M $-$ A by whether the prompt changes (post hoc).} The A/M equality was checked by rerendering all 360 prompts. Differences average state-level rates from valid replies; 90\% catalog-by-model stratified bootstrap CIs, with a run-then-state hierarchical sensitivity for the changed subset. Per-model rows subdivide that subset. The registered H15 equivalence test remains the all-state contrast in Table~\ref{tab:r4}.}
\label{tab:history-sensitivity}
\end{table*}

\begin{table}[h]
\centering\scriptsize\setlength{\tabcolsep}{2pt}
\resizebox{\columnwidth}{!}{\begin{tabular}{@{}llll@{}}
\toprule
Contrast & Gemma-4-26B-A4B & Mistral-Small-3.2-24B & Qwen3-VL-32B \\
\midrule
B $-$ A & $+$0.03 [$-$0.05, 0.12] & $-$0.31 [$-$0.45, $-$0.17] & $-$0.22 [$-$0.30, $-$0.13] \\
R $-$ A & $+$0.05 [$-$0.03, 0.14] & $-$0.04 [$-$0.18, 0.10] & $-$0.21 [$-$0.30, $-$0.12] \\
J $-$ K & $+$0.17 [0.05, 0.30] & $+$0.04 [$-$0.17, 0.24] & $-$0.02 [$-$0.16, 0.12] \\
A $-$ H & $-$0.57 [$-$0.71, $-$0.42] & $+$0.31 [0.12, 0.49] & $-$0.36 [$-$0.53, $-$0.19] \\
F $-$ N & $-$0.14 [$-$0.25, $-$0.04] & $-$0.22 [$-$0.37, $-$0.05] & $-$0.11 [$-$0.18, $-$0.05] \\
P $-$ F & $+$0.00 [$-$0.07, 0.08] & $-$0.15 [$-$0.26, $-$0.05] & $-$0.01 [$-$0.04, 0.00] \\
E $-$ A & $-$0.05 [$-$0.09, 0.00] & $-$0.53 [$-$0.64, $-$0.41] & $-$0.24 [$-$0.33, $-$0.16] \\
\bottomrule
\end{tabular}
}
\caption{Round-4 state-level rate differences per model (fresh states; registered as descriptive, no multiplicity correction), with bootstrap CIs over states. The models differ in kind: deletion and content effects are small for Gemma-4, deletion acts mostly through non-semantic text bulk for Mistral, and the history raises re-proposal for Mistral only; the status sentence (F $-$ N) is negative for all three.}
\label{tab:r4models}
\end{table}

\begin{table*}[t]
\centering\scriptsize
\resizebox{\textwidth}{!}{\begin{tabular}{@{}lllll@{}}
\toprule
Proposer (states) & $B-A$ & $C-A$ & $D-A$ & $E-A$ \\
\midrule
Gemma-4-26B-A4B (37) & $-$0.11 [$-$0.19, $-$0.05] (0.054) & $-$0.01 [$-$0.12, 0.09] (1.000) & $-$0.19 [$-$0.31, $-$0.09] (0.018) & $-$0.16 [$-$0.27, $-$0.05] (0.063) \\
Mistral-Small-3.2-24B (40) & $-$0.31 [$-$0.45, $-$0.17] (0.002) & $+$0.03 [$-$0.09, 0.15] (1.000) & $-$0.11 [$-$0.23, 0.01] (0.546) & $-$0.57 [$-$0.69, $-$0.43] ($<$0.001) \\
Qwen3-VL-32B (40) & $-$0.14 [$-$0.22, $-$0.06] (0.021) & $+$0.10 [0.00, 0.20] (0.546) & $-$0.05 [$-$0.15, 0.06] (1.000) & $-$0.16 [$-$0.23, $-$0.10] ($<$0.001) \\
\bottomrule
\end{tabular}
}
\caption{Per-model state-level rate differences on the original 120-state intervention round (registered secondary family; Holm correction over 12 contrasts; $p_\mathrm{Holm}$ in parentheses). States with no valid reply under A are dropped from that model's contrasts.}
\label{tab:permodel}
\end{table*}

\begin{table*}[t]
\centering\scriptsize
\begin{tabular}{@{}lllrrrrrr@{}}
\toprule
Proposer & Cat. & View & Seed & Elig. & Valid & $B$ & Kept & Final dev \\
\midrule
Gemma-4-26B-A4B & A & FULL & 20 & 26 & 21 & 5 & 3 & 0.850 \\
Gemma-4-26B-A4B & A & FULL & 21 & 27 & 17 & 1 & 3 & 0.867 \\
Gemma-4-26B-A4B & A & FULL & 22 & 25 & 23 & 8 & 3 & 0.850 \\
Gemma-4-26B-A4B & A & W8 & 20 & 25 & 22 & 2 & 3 & 0.850 \\
Gemma-4-26B-A4B & A & W8 & 21 & 26 & 24 & 1 & 3 & 0.850 \\
Gemma-4-26B-A4B & A & W8 & 22 & 25 & 23 & 5 & 3 & 0.867 \\
Gemma-4-26B-A4B & B & FULL & 20 & 28 & 27 & 8 & 1 & 0.800 \\
Gemma-4-26B-A4B & B & FULL & 21 & 28 & 26 & 5 & 1 & 0.800 \\
Gemma-4-26B-A4B & B & FULL & 22 & 27 & 26 & 11 & 2 & 0.817 \\
Gemma-4-26B-A4B & B & W8 & 20 & 28 & 26 & 5 & 1 & 0.800 \\
Gemma-4-26B-A4B & B & W8 & 21 & 27 & 26 & 3 & 2 & 0.817 \\
Gemma-4-26B-A4B & B & W8 & 22 & 27 & 25 & 2 & 2 & 0.817 \\
Mistral-Small-3.2-24B & A & FULL & 20 & 28 & 28 & 21 & 1 & 0.800 \\
Mistral-Small-3.2-24B & A & FULL & 21 & 27 & 27 & 19 & 2 & 0.817 \\
Mistral-Small-3.2-24B & A & FULL & 22 & 29 & 29 & 23 & 0 & 0.750 \\
Mistral-Small-3.2-24B & A & W8 & 20 & 27 & 27 & 19 & 2 & 0.850 \\
Mistral-Small-3.2-24B & A & W8 & 21 & 28 & 28 & 18 & 1 & 0.783 \\
Mistral-Small-3.2-24B & A & W8 & 22 & 28 & 28 & 21 & 1 & 0.800 \\
Mistral-Small-3.2-24B & B & FULL & 20 & 27 & 27 & 20 & 2 & 0.817 \\
Mistral-Small-3.2-24B & B & FULL & 21 & 27 & 27 & 20 & 2 & 0.833 \\
Mistral-Small-3.2-24B & B & FULL & 22 & 27 & 27 & 13 & 2 & 0.817 \\
Mistral-Small-3.2-24B & B & W8 & 20 & 28 & 28 & 20 & 1 & 0.783 \\
Mistral-Small-3.2-24B & B & W8 & 21 & 26 & 25 & 14 & 3 & 0.850 \\
Mistral-Small-3.2-24B & B & W8 & 22 & 26 & 24 & 7 & 3 & 0.850 \\
Qwen3-VL-32B & A & FULL & 20 & 27 & 27 & 1 & 2 & 0.817 \\
Qwen3-VL-32B & A & FULL & 21 & 27 & 27 & 6 & 2 & 0.817 \\
Qwen3-VL-32B & A & FULL & 22 & 26 & 26 & 3 & 3 & 0.867 \\
Qwen3-VL-32B & A & W8 & 20 & 27 & 27 & 4 & 2 & 0.817 \\
Qwen3-VL-32B & A & W8 & 21 & 27 & 27 & 1 & 2 & 0.817 \\
Qwen3-VL-32B & A & W8 & 22 & 27 & 27 & 1 & 2 & 0.817 \\
Qwen3-VL-32B & B & FULL & 20 & 27 & 25 & 8 & 2 & 0.817 \\
Qwen3-VL-32B & B & FULL & 21 & 28 & 28 & 16 & 1 & 0.800 \\
Qwen3-VL-32B & B & FULL & 22 & 27 & 22 & 6 & 2 & 0.817 \\
Qwen3-VL-32B & B & W8 & 20 & 28 & 28 & 9 & 1 & 0.800 \\
Qwen3-VL-32B & B & W8 & 21 & 26 & 24 & 0 & 3 & 0.850 \\
Qwen3-VL-32B & B & W8 & 22 & 26 & 25 & 3 & 3 & 0.850 \\
\bottomrule
\end{tabular}

\caption{Every replication loop. Elig.: steps with $X_t\neq\varnothing$; Valid: those with a valid reply; $B$: re-proposals of a visible rejection; Kept: edits in the final harness; Final dev: final development accuracy.}
\label{tab:runs}
\end{table*}

\section{The Two Catalogs}\label{app:catalogs}

Table~\ref{tab:catalog} lists the 40 edits with their catalog-A and catalog-B IDs. Catalog-B descriptions are reworded; for example, E03 ``Accuracy is very important for this task.'' becomes ``Add to the agent's instructions: getting this right matters a lot.'' Catalog~B groups the options under headings (prompt instructions, rendering, decoding), and its task text, question and history line format differ from catalog~A.

\begin{table*}[t]
\centering\scriptsize
\begin{tabular}{@{}lllp{0.62\textwidth}@{}}
\toprule
A & B & Kind & Catalog-A description \\
\midrule
E01 & RF1 & prompt & Briefly describe the screen before choosing an action. \\
E02 & ZQ7 & prompt & Double-check your answer before replying. \\
E03 & NK3 & prompt & Accuracy is very important for this task. \\
E04 & WV1 & decode & Lower the output token limit from 48 to 16 to save cost. \\
E05 & VG3 & prompt & If you are unsure, choose the most prominent element on the... \\
E06 & YV2 & prompt & Put the final reply on the last line, prefixed with FINAL:. \\
E07 & AF9 & render & Enlarge the tag labels to 13px bold. \\
E08 & FG6 & prompt & Consider what the user is ultimately trying to accomplish. \\
E09 & DG4 & prompt & In label-value rows the value is right-aligned on the same ... \\
E10 & YJ8 & render & Outline every tagged element with a thin box in the tag's c... \\
E11 & BK6 & prompt & Calendar events are coloured blocks whose label starts with... \\
E12 & AN3 & prompt & In the inbox, the sender's name is the bold first line of e... \\
E13 & WN6 & prompt & For questions, copy the value exactly as displayed, includi... \\
E14 & NM3 & render & Increase screenshot contrast by 30\%. \\
E15 & PZ9 & prompt & Take a deep breath and work on this problem step by step. \\
E16 & VK8 & prompt & Icon legend: [icon] back, [icon] archive, [icon] delete, [i... \\
E17 & CV5 & render & Hide the tags of elements that are covered by another eleme... \\
E18 & GZ3 & prompt & Switches: green with the knob on the right means On; grey w... \\
E19 & MV5 & decode & Raise the output token limit from 48 to 256 so the agent ca... \\
E20 & CD7 & prompt & If a dialog or pop-up is open, only the elements inside the... \\
E21 & RT8 & prompt & To toggle a setting, tap its switch rather than the text la... \\
E22 & JV4 & prompt & Be careful and precise. \\
E23 & DZ8 & prompt & First list the tags you can see, then choose one. \\
E24 & RX7 & prompt & Here is an example of a correct reply to 'Open settings': T... \\
E25 & JJ6 & prompt & Do not explain your answer; output only the single reply line. \\
E26 & HW7 & prompt & Each tag number is drawn just ABOVE the top-left corner of ... \\
E27 & CM1 & render & Downscale screenshots to 0.75x to save visual tokens. \\
E28 & XH2 & render & Render screenshots at 1.5x resolution (more visual tokens). \\
E29 & RA1 & prompt & You are an expert mobile GUI agent with years of experience. \\
E30 & XR4 & render & Draw each tag inside the top-left corner of its element ins... \\
E31 & KR8 & prompt & Tasks that ask you to do something (open, turn on, confirm,... \\
E32 & HR6 & prompt & Prefer elements in the top toolbar when several elements co... \\
E33 & JG7 & prompt & Tags are numbered in reading order: top to bottom, left to ... \\
E34 & WA9 & prompt & Answer confidently; do not hedge. \\
E35 & QD8 & prompt & Greyed-out buttons are disabled and cannot be pressed. \\
E36 & ZJ8 & render & Convert screenshots to grayscale. \\
E37 & PM2 & prompt & Reply as JSON: \{"action": "TAP" or "ANSWER", "value": \textless{}tag ... \\
E38 & XX3 & prompt & Ordinal words in a task (first, fifth, ...) count items on ... \\
E39 & AA3 & prompt & Think step by step before giving your final answer. \\
E40 & BW8 & render & Give each tag a distinct colour instead of all red. \\
\bottomrule
\end{tabular}

\caption{The 40 edits. A: catalog-A ID; B: catalog-B code; kind; catalog-A description (truncated at 62 characters).}
\label{tab:catalog}
\end{table*}

\section{Codebook}\label{app:codebook}

The codebook (\texttt{notes/CODEBOOK.md}) defines eight registered codes: explicit retry, failure-target attraction, rejection reinterpretation, history neglect, salience copying, explicit exclusion checking, ID/identity confusion and other. Its decision rules were added by annotator~1 after reading the 60 sampled replies, before recording any code: (i) codes 1 and 3 concern the chosen edit; (ii) code 6 applies when the reply names an earlier attempt that was not kept and uses it as a reason to choose something else, or says that it must pick something untried; (iii) code 2 applies if the rationale cites the shown failures, and code 4 if it argues only from the catalog or from general principles; (iv) code 7 includes treating an edit not in the harness as if it were. Precedence is 1, 3, 6, 7, then the dominant of 2, 4 and 5. Blinded DeepSeek agrees with annotator~1 at $\kappa=$\KappaAD, and all three annotators agree on \AllThree{} of 60 replies, \ConsFTA{} of them failure-target attraction.

\section{Positioning Matrix}\label{app:matrix}

\begin{table*}[t]
\centering\small
\begin{tabularx}{\textwidth}{@{}>{\raggedright\arraybackslash}p{4.3cm}>{\raggedright\arraybackslash}X>{\raggedright\arraybackslash}X>{\raggedright\arraybackslash}X@{}}
\toprule
Line of work & What is established & Assumption we test & How we test it \\
\midrule
LLM optimizers with shown history~\citep{yang2024opro,agrawal2026gepa,lee2026metaharness,lu2026procgraphs,yang2026skillopt} & showing tried candidates and failures helps search & a shown rejection is avoided & $\alpha$ in reached states; no-history baseline (\S\ref{sec:res-transport}) \\
Static instruction following~\citep{zhou2023ifeval,jiang2024followbench} & compliance in one response & avoidance is a trait measured once & clean probe vs.\ own loop vs.\ reconstructed state (\S\ref{sec:res-transport}) \\
Negative and omission constraints~\citep{castricato2024pink,rana2026gravity,gamage2026omission} & prohibitions are fragile; prior pressure drives violations & pressure is a fixed prior & no-history baseline at the reached harness; status vs.\ prohibition (\S\ref{sec:res-signal}) \\
Failure records and repetition~\citep{gumaan2026backfires,yang2026retrying,xu2022repetition} & a visible failed action is repeated; ``do not repeat'' does not help (135M--1.7B) & the failure record drives repetition & approximately length-matched neutral text, attachment shuffle, status sentence (\S\ref{sec:res-failure}--\ref{sec:res-signal}) \\
Knowing vs.\ acting~\citep{kruthof2026drift,dong2026memarbiter} & models restate what they violate & stated reasons explain choices & stated reasons vs.\ content controls (\S\ref{sec:res-boundary}) \\
Reasoning vs.\ compliance~\citep{fu2025mathif,li2025thinkingfails} & reasoning lowers static-instruction compliance & this generalises to all constraints & instruct/thinking pair, provider-pinned (\S\ref{sec:res-boundary}) \\
\bottomrule
\end{tabularx}
\caption{\textbf{Positioning matrix.} Each line of work, what it establishes, and the assumption that a history-implied exclusion constraint puts under test.}
\label{tab:matrix}
\end{table*}

Table~\ref{tab:matrix} summarises how each line of related work (\S\ref{sec:related}) relates to the question of this paper.

\section{Earlier Evidence on the Same Benchmark}\label{app:earlier}

An earlier study on this benchmark and catalog~A (unpublished; its data are read-only inputs here) motivated the present design. Its results are not part of this paper's registration, and we list them with their original status.

\paragraph{Probe and loop (exploratory).} Four open proposers declined their own rejected favourite at 81--100\% of probe positions, but declined visible rejections in only 28--82\% of loop steps.

\paragraph{Near-miss control (exploratory).} Over (visible rejection, step) pairs, clearly failed edits (development change $\le-2$ items) were re-proposed in \NMfailB{} of \NMfailN{} pairs (\NMfail), and near-misses (within one item of the threshold) in \NMnearB{} of \NMnearN{} (\NMnear). Within runs neither was consistently higher (near-miss higher in \NMnearHi{}, failed higher in \NMfailHi{} of \NMruns{} runs), so a rational retry of near-misses does not explain re-proposal.

\paragraph{Instructed loops (earlier registered).} A true sentence stating that evaluation is deterministic and that rejected edits must not be proposed again changed the pooled rate from \UNrate{} (\UNb/\UNn) to \INSrate{} (\INSb/\INSn), in runs compared with uninstructed runs on the same seeds.

\paragraph{Replay on violation-selected states (earlier registered; biased).} In \RPn{} stored states where a re-proposal had occurred, a single replay call reproduced a visible re-proposal in \RPctl. Removing failure examples gave \RPnofail{} and a DO-NOT list gave \RPdnp{} (Table~\ref{tab:replayold}). Selecting states on a prior violation biases these rates upward and makes removal effects partly regression to the mean; one call per state cannot separate an intervention from temperature-0 variation. The random-state design of \S\ref{sec:design-transport} replaces this analysis.

\begin{table}[h]
\centering\scriptsize\setlength{\tabcolsep}{3pt}
\begin{tabular}{@{}lrrrr@{}}
\toprule
Proposer & States & Replay (A) & No failures (B) & DO-NOT (E) \\
\midrule
Gemma-4-26B-A4B & 27 & 10 & 2 & 2 \\
Mistral-Small-3.2-24B & 48 & 39 & 42 & 6 \\
Qwen3-VL-32B & 44 & 18 & 0 & 0 \\
\bottomrule
\end{tabular}

\caption{Earlier replay on violation-selected states (one call per state): states with a visible re-proposal.}
\label{tab:replayold}
\end{table}

\paragraph{Reasoning-mode proposer (earlier registered).} gpt-oss-120b~\citep{openai2025gptoss} honoured \AthreeProbe{} probe cells and re-proposed in \AthreeLoopB{} of \AthreeLoopN{} loop steps (rate \AthreeLoopRate).

\end{document}